%% file: neurips_2026.tex
\documentclass{article}

\usepackage[preprint]{neurips_2026}

\usepackage[utf8]{inputenc} 
\usepackage[T1]{fontenc}    
\usepackage{hyperref}       
\usepackage{url}            
\usepackage{booktabs}       
\usepackage{amsfonts}       
\usepackage{nicefrac}       
\usepackage{microtype}      
\usepackage{xcolor}         
\usepackage{amsmath}
\usepackage{graphicx}
\usepackage{adjustbox}
\graphicspath{{figures/}}
\title{NSVQ: Mitigating Codebook Collapse by Stabilizing Encoder Drift in Vector Quantization}

\author{%
  Hao Lu \\
  Wake Forest University School of Medicine / Advocate Health \\
  \texttt{hao.lu@advocatehealth.org} \\
  \And
  Yongxin Guo \\
  Wake Forest University School of Medicine \\
  \texttt{Yongxin.Guo@wfusm.edu} \\
  \And
  Onur Can Koyun \\
  Advocate Health \\
  \texttt{Onur.Koyun@advocatehealth.org} \\
  \And
  Zhengjie Zhu \\
  Wake Forest University School of Medicine \\
  \texttt{Zhengjie.Zhu@wfusm.edu} \\
  \And
  Abbas Alili \\
  Advocate Health \\
  \texttt{Abbas.Alili@advocatehealth.org} \\
  \And
  Metin Nafi Gurcan \\
  Wake Forest University School of Medicine \\
  \texttt{Metin.Gurcan@wfusm.edu} \\
}

\begin{document}

\maketitle

\input{main}
\clearpage
{
    \small
    \bibliographystyle{plainnat}
    \bibliography{main}
}
\clearpage
\input{appendix}

\end{document}

%% file: main.tex
\begin{abstract}
Vector quantization is central to modern generative modeling pipelines, but large-codebook VQ models often suffer from codebook collapse. We identify encoder drift as a key driver of this failure: as the encoder moves the latent distribution, sparsely updated code vectors can lag behind, lose assignments, and increase quantization error, creating a feedback loop through the straight-through estimator. We propose NSVQ, a non-stationary-aware VQ training strategy that combines a dense non-stationary embedding loss, codebook replacement, and stage-wise encoder freezing. NSVQ first helps the codebook track encoder drift during early training, then freezes the encoder to consolidate the codebook under a fixed latent geometry, and finally reintroduces adversarial refinement. Experiments on ImageNet-1k show that NSVQ improves reconstruction quality while maintaining full codebook utilization. On ImageNet-1k at 128$\times$128 with 65,536 codes, NSVQ reduces rFID from 2.39 to 2.10 compared with SimVQ, while both methods maintain 100\% utilization. Additional latent diffusion experiments show that NSVQ also improves downstream ImageNet generation FID.
\end{abstract}


\section{Introduction}

Vector quantization (VQ) represents continuous data with discrete codes~\citep{gray1998quantization}. Since VQ-VAE~\citep{oord2017neural}, it has become a core component of modern generative modeling systems, including VQGAN-style tokenizers~\citep{esser2021taming}, latent diffusion models~\citep{rombach2022high}, and discrete visual tokenizers for vision-language modeling~\citep{bao2022beit}. In these systems, images are compressed into discrete token sequences for autoregressive, masked-prediction, or multimodal modeling, making quantizer stability and reconstruction quality central to downstream performance.

A persistent challenge in VQ training is codebook collapse. As the codebook grows, many codes may remain inactive, reducing the effective capacity of the discrete representation. Existing methods improve utilization through stochastic quantization, regularization, code replacement, or mapping-based codebook parameterizations, but they do not fully explain why codes lose assignments or why high utilization alone may still fail to yield a high-quality tokenizer. In this work, we view these issues as consequences of a non-stationary alignment problem driven by encoder drift.

During VQ-VAE training, the encoder is continuously updated through the straight-through estimator (STE)~\citep{bengio2013estimating}, causing the latent distribution to move, whereas the codebook is updated sparsely through nearest-neighbor assignments. Code vectors must therefore serve as stable discrete representatives for a latent space that is itself drifting. When this mismatch becomes severe, codes may lose assignments and collapse; even when utilization remains high, drift-induced misalignment can degrade reconstruction and downstream generation by weakening consistency among the encoder, codebook, and decoder.

Motivated by this view, we propose NSVQ, a non-stationary-aware VQ training strategy aimed at improving codebook quality, not merely codebook usage. NSVQ first compensates for encoder-induced non-stationarity by propagating local auxiliary updates to nearby non-winning codes, helping the codebook track the moving latent geometry. It then freezes the encoder, performs a short quantizer--decoder warm-up to consolidate the codebook, and finally reintroduces adversarial refinement under a fixed latent geometry. Thus, NSVQ treats collapse as one symptom of non-stationary misalignment while targeting stable encoder--codebook--decoder coordination.

Our contributions are threefold. First, we identify encoder drift as a source of non-stationary encoder--codebook misalignment in large-codebook VQ training, with codebook collapse as an extreme consequence. Second, we introduce a drift-aware codebook update that improves local alignment by propagating auxiliary updates to nearby non-winning codes while preserving hard nearest-neighbor quantization. Third, we propose a stage-wise schedule that compensates for non-stationarity, freezes the encoder for codebook consolidation, and performs adversarial refinement without reintroducing encoder drift. Empirically, on ImageNet-1k at $128\times128$ with 65,536 codes, NSVQ reduces rFID from 2.39 to 2.10 compared with SimVQ while maintaining 100\% utilization, and also improves downstream latent-diffusion generation under the same U-Net setting.

\section{Background and Problem Formulation}

\subsection{Preliminaries on VQ-VAE}

A VQ-VAE consists of an encoder $E_\theta$, a decoder $D_\phi$, and a codebook
$C=\{c_1,\ldots,c_K\}$~\citep{oord2017neural}. Given an input image $x$, the encoder produces
$z_e=E_\theta(x)$, which is quantized by nearest-neighbor assignment:
\begin{equation}
q=\arg\min_j \|z_e-c_j\|_2^2,\qquad z_q=c_q,\qquad \hat{x}=D_\phi(z_q).
\end{equation}
The standard objective combines reconstruction, embedding, and commitment losses:
\begin{equation}
\mathcal{L}
=
\mathcal{L}_{\rm rec}
+
\|\mathrm{sg}[z_e]-z_q\|_2^2
+
\beta\|z_e-\mathrm{sg}[z_q]\|_2^2 ,
\end{equation}
where $\mathrm{sg}[\cdot]$ denotes stop-gradient. The STE allows decoder gradients to update the encoder through the non-differentiable quantizer, while
the codebook itself is updated only through assignment-based embedding losses.

\subsection{Encoder Drift and the Quantization-Error Feedback Loop}
\label{sec:background_encoder_drift}

The key issue studied in this paper is that the encoder distribution is non-stationary during VQ
training. We define \emph{encoder drift} as the change in encoder outputs for the same input over
training:
\begin{equation}
z_e^{(t)}(x)=E_{\theta^{(t)}}(x).
\end{equation}
Because $\theta^{(t)}$ is continuously updated by STE gradients, the latent distribution can move
throughout training. In contrast, codebook updates are sparse: for a mini-batch at step $t$, only
codes selected by nearest-neighbor assignment receive direct embedding updates,
\begin{equation}
c_j^{(t+1)}
=
c_j^{(t)}
+
\eta_c
\sum_{i:q_i=j}
\left(z_{e,i}^{(t)}-c_j^{(t)}\right).
\end{equation}
Non-selected codes therefore do not directly follow the moving latent geometry. This creates a
non-stationary alignment problem: encoder drift can make code vectors lag behind, lose assignments,
and become inactive.

This process can further amplify itself through quantization error. Write
\begin{equation}
z_q = z_e + \epsilon,
\qquad
\epsilon = z_q - z_e ,
\end{equation}
where $\epsilon$ is the residual induced by replacing a continuous latent point with a discrete code
vector. As codebook lag and dead codes accumulate, active codes become poorer representatives of
the encoder distribution, increasing $\|\epsilon\|$. Since the STE passes decoder gradients to the
encoder as if quantization were identity-like, the encoder receives gradients evaluated at $z_q$ rather
than at $z_e$. If $\ell(z)$ denotes the decoder-side loss, a local first-order approximation gives
\begin{equation}
g_{\theta}^{\rm STE} - g_{\theta}^{\rm cont}
\approx
J_{\theta}(x)^\top H_{\ell}(z_e)\epsilon,
\end{equation}
where $J_{\theta}(x)=\partial E_{\theta}(x)/\partial \theta$ and $H_{\ell}(z_e)$ is the local Hessian of
the decoder-side loss. Thus, larger quantization residuals can enlarge the gap between the STE
gradient and the continuous-latent gradient.

This yields a positive feedback loop:
\begin{equation*}
\text{encoder drift}
\rightarrow
\text{codebook lag}
\rightarrow
\text{dead codes}
\rightarrow
\|\epsilon\| \uparrow
\rightarrow
\text{STE gradient-estimation gap}
\rightarrow
\text{encoder drift} \uparrow .
\end{equation*}
Under this view, codebook collapse is not only the consequence of sparse updates; it can become
part of a self-amplifying instability that eventually causes catastrophic reconstruction degradation.
Additional controlled evidence beyond poor initialization is provided in Appendix~\ref{app:controlled-drift}.

For empirical analysis, we measure encoder drift by the root-mean-square change of encoder outputs
on a fixed evaluation subset $S$:
\begin{equation}
D_{\rm RMS}^{t}
=
\sqrt{
\mathbb{E}_{x\in S,u}
\left[
\frac{1}{d}
\left\|
E_{\theta_t}(x)_u - E_{\theta_{t-\Delta}}(x)_u
\right\|_2^2
\right]
}.
\label{eq:encoder-drift}
\end{equation}
Here, $u$ indexes spatial latent positions, $d$ is the latent channel dimension, and $\Delta$ is the
checkpoint logging interval. Since $S$ is fixed, changes in $D_{\rm RMS}^{t}$ reflect encoder movement
rather than changes in sampled inputs.
\section{Related Work}
\label{sec:related-work}

Existing approaches to codebook collapse can be grouped into heuristic methods, mapping-based codebook adaptation, and frozen-representation tokenization.

\paragraph{Heuristic collapse mitigation.}
Soft or stochastic quantization methods~\citep{roy2018theory,williams2020hierarchical,takida2022sqvae} relax hard nearest-neighbor assignment so that more codes receive training signal. Regularized VQ~\citep{zhang2023regularized} and VQGAN-FC~\citep{yu2022vector} improve code usage through regularization or architectural modifications. Another practical strategy replaces inactive codes with active latent features, as used in Jukebox~\citep{dhariwal2020jukebox}, SoundStream~\citep{zeghidour2021soundstream}, and Online Clustered Codebook~\citep{zheng2023online}. These methods improve utilization, but mainly treat collapse as an assignment or code-maintenance problem. NSVQ instead views collapse as a non-stationary optimization problem caused by encoder drift and sparse codebook updates.

\paragraph{Mapping-based dense codebook adaptation.}
Affine codebook reparameterization~\citep{huh2023straightening}, VQGAN-LC~\citep{zhu2024scaling}, SimVQ~\citep{zhu2024addressing}, and FVQ~\citep{chang2025scalable} reduce collapse by parameterizing the effective codebook through shared transformations. Updating the mapping can move many codes simultaneously, alleviating the sparse-update limitation of standard VQ. However, the movement of non-winning codes is implicit and not directly tied to measured encoder drift. NSVQ is complementary: it explicitly propagates local auxiliary updates to nearby non-winning codes during the non-stationary phase of training.

\paragraph{Frozen-representation tokenization.}
Recent tokenizers freeze pretrained or matured visual representations and optimize a quantizer, decoder, or rectifier around them, including CODA~\citep{liu2025coda}, ReVQ~\citep{zhang2025revq}, VQ-Transplant~\citep{fang2026vqtransplant}, and QLIP~\citep{zhao2025qlip}. These works focus on representation reuse or continuous-to-discrete conversion, rather than encoder drift as a cause of collapse. NSVQ also freezes the encoder, but only after joint VQ training has shaped a reconstruction-oriented and quantization-aware latent space. This timing matters because representation learning and clustering can reinforce each other, as shown by Deep Embedded K-Means Clustering~\citep{guo2021deep}; a frozen vision representation is not necessarily cluster-structured or quantization-friendly. Thus, NSVQ uses freezing as late-stage consolidation to remove further latent drift, not as an initial tokenization assumption.

\section{Proposed NSVQ Training Strategy}
\label{sec:method}

\begin{figure}[t]
\centering
\includegraphics[width=\linewidth]{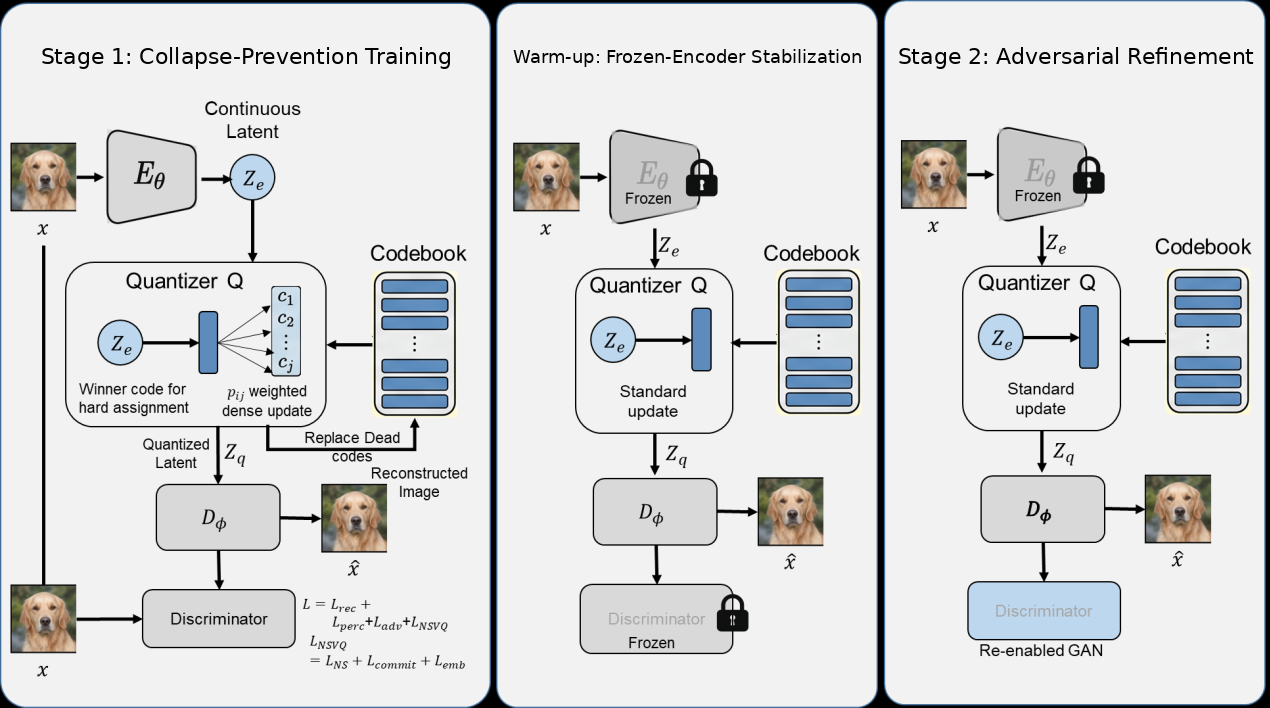}
\caption{Overview of NSVQ. Stage~1 prevents early collapse using the non-stationary embedding loss and codebook replacement while the encoder is still drifting. The intermediate warm-up stage freezes the encoder and disables adversarial training, allowing the quantizer and decoder to adapt to a fixed latent geometry without VQGAN instability. Stage~2 re-enables adversarial training for perceptual refinement while keeping the encoder frozen.}
\label{fig:overview}
\end{figure}

NSVQ stabilizes large-codebook VQ training by matching the training procedure to the encoder-drift dynamics identified in Section~\ref{sec:background_encoder_drift}. Early in training, encoder drift is strong, so NSVQ first compensates for the moving latent geometry, then freezes the encoder once the encoder--codebook interaction has stabilized, and finally performs adversarial refinement under a fixed latent geometry. The overall procedure is shown in Figure~\ref{fig:overview}.

\subsection{Stage 1: Non-Stationary-Aware Codebook Learning}

Stage~1 corresponds to the highly non-stationary phase of VQ training. The encoder, decoder,
quantizer, and discriminator are trained jointly. For each latent vector $z_i$, we compute its squared
Euclidean distance to each code vector:
\begin{equation}
d_{ij}
=
\left\|
\operatorname{sg}[z_i]-c_j
\right\|_2^2,
\qquad
q_i=\arg\min_j d_{ij}.
\end{equation}
Here, $q_i$ denotes the winning code index used for the forward quantized representation. Standard
VQ updates only this winning code through the embedding loss. Under encoder drift, however, nearby
non-winning codes may also need to follow the moving latent distribution; otherwise, they can lag
behind and eventually lose future assignments.

To propagate drift-aware training signal beyond the winning code, NSVQ assigns temperature-controlled
soft weights to all codes:
\begin{equation}
p_{ij}
=
\frac{\exp(-d_{ij}/\tau)}
{\sum_{k=1}^{K}\exp(-d_{ik}/\tau)},
\qquad
\tilde p_{ij}=\operatorname{sg}[p_{ij}] .
\label{eq:ns-soft-weight}
\end{equation}
As detailed in Appendix~\ref{app:nsvq-derivation}, the auxiliary direction can be interpreted as
combining a local encoder-codebook residual with a catch-up correction toward the selected active
code, yielding the simple attractive direction $\operatorname{sg}[z_i]-c_j$ for each non-winning code.
The non-stationary embedding loss is then defined as
\begin{equation}
\mathcal{L}_{NS}
=
\frac{1}{Nd}
\sum_i
\sum_{j\ne q_i}
\tilde p_{ij}d_{ij}.
\label{eq:ns-loss}
\end{equation}
where $N$ is the number of latent positions and $d$ is the code dimension. The detached weights
$\tilde p_{ij}$ determine the relative strength of the auxiliary updates, giving stronger updates to
codes close to the current latent point and weaker updates to distant codes. The winning code is
excluded from $\mathcal{L}_{NS}$ because it is already updated by the standard embedding loss; its
probability mass is not redistributed to non-winning codes.

Importantly, $\mathcal{L}_{NS}$ does not replace hard nearest-neighbor quantization with soft
quantization. The forward pass still uses the discrete code $c_{q_i}$, so the VQ bottleneck remains
unchanged. The soft weights are used only to distribute auxiliary embedding updates to nearby
non-winning codes. From the feedback-loop view in Section~\ref{sec:background_encoder_drift},
the role of $\mathcal{L}_{NS}$ is therefore not merely to increase utilization. By moving nearby
non-winning codes toward the current encoder outputs, the codebook can better track the drifting
latent geometry before assignments are lost. This limits the growth of the quantization residual
$\epsilon=z_q-z_e$, thereby weakening the STE-induced feedback from quantization error back to
encoder drift.

The Stage-1 quantization loss is
\begin{equation}
\mathcal{L}_{\mathrm{vq}}^{(1)}
=
\left\|
z_q-\operatorname{sg}[z_e]
\right\|_2^2
+
\beta
\left\|
z_e-\operatorname{sg}[z_q]
\right\|_2^2
+
\alpha
\mathcal{L}_{NS},
\end{equation}
where $\beta$ controls the commitment loss and $\alpha$ controls the strength of the non-stationary
correction. The full generator objective is
\begin{equation}
\mathcal{L}^{(1)}
=
\mathcal{L}_{\mathrm{rec}}
+
\lambda_{\mathrm{perc}}\mathcal{L}_{\mathrm{perc}}
+
\lambda_{\mathrm{adv}}\mathcal{L}_{\mathrm{adv}}
+
\lambda_{\mathrm{vq}}\mathcal{L}_{\mathrm{vq}}^{(1)}.
\end{equation}

Stage~1 also uses codebook replacement to revive persistently inactive entries. Replacement is used
only during Stage~1, where encoder drift is strongest and dead-code recovery is most important. In
this sense, the dense non-stationary loss and replacement mechanism play complementary roles:
$\mathcal{L}_{NS}$ continuously reduces codebook lag for nearby codes, while replacement prevents
already inactive codes from remaining permanently unused. Together, this Stage-1 protection helps
break the feedback loop between encoder drift, dead codes, quantization error, and further encoder
drift.

\paragraph{Encoder-freezing criterion.}
We use the epoch-averaged commitment loss as a practical indicator of whether the encoder--codebook
interaction has stabilized. Let $\ell_t^{\mathrm{commit}}$ denote the average commitment loss over
epoch $t$:
\begin{equation}
\ell_t^{\mathrm{commit}}
=
\frac{1}{|\mathcal{B}_t|}
\sum_{b\in\mathcal{B}_t}
\mathcal{L}_{\mathrm{commit}}(b),
\end{equation}
where $\mathcal{B}_t$ is the set of mini-batches in epoch $t$, and
\begin{equation}
\mathcal{L}_{\mathrm{commit}}
=
\left\|
z_e-\operatorname{sg}[z_q]
\right\|_2^2.
\end{equation}
The encoder is frozen when the epoch-averaged commitment loss has not decreased for 10 consecutive
epochs. Equivalently, if no new lower value of $\ell_t^{\mathrm{commit}}$ is observed within a
10-epoch patience window, we regard the encoder--codebook interaction as having reached a plateau
and enter the frozen-encoder warm-up stage.

Intuitively, a decreasing commitment loss indicates that the encoder outputs and the selected code
vectors are still adapting to each other. In contrast, if the epoch-averaged commitment loss fails to
decrease for 10 consecutive epochs, the latent geometry has largely stabilized. At this point, further
encoder updates provide limited benefit but can continue to introduce non-stationarity into codebook
learning. NSVQ therefore freezes the encoder and enters the intermediate warm-up stage before
adversarial refinement.

\subsection{Warm-up: Frozen-Encoder Stabilization}

After the commitment-loss plateau criterion is met, NSVQ enters a short frozen-encoder warm-up stage. The encoder is frozen, while adversarial training, $\mathcal{L}_{\mathrm{NS}}$, and codebook replacement are disabled. The purpose of this stage is to prevent VQGAN collapse immediately after encoder freezing. Although the latent distribution is now stationary, the quantizer and decoder still need to adapt to the fixed encoder manifold. Removing GAN pressure during this transition allows them to stabilize with reconstruction-oriented losses before adversarial refinement is reintroduced.

The warm-up quantization loss is reduced to a codebook-fitting objective:
\begin{equation}
\mathcal{L}_{\mathrm{vq}}^{(\mathrm{warm})}
=
\left\|
z_q-\operatorname{sg}[z_e]
\right\|_2^2.
\end{equation}
The warm-up objective is
\begin{equation}
\mathcal{L}^{(\mathrm{warm})}
=
\mathcal{L}_{\mathrm{rec}}
+
\lambda_{\mathrm{perc}}\mathcal{L}_{\mathrm{perc}}
+
\lambda_{\mathrm{vq}}\mathcal{L}_{\mathrm{vq}}^{(\mathrm{warm})}.
\end{equation}
Thus, the warm-up stage acts as a collapse-safe transition between non-stationary codebook learning and adversarial perceptual refinement.

\subsection{Stage 2: Perceptual Refinement under Fixed Latent Geometry}

Stage~2 keeps the encoder frozen and reintroduces adversarial training. The quantization loss remains the same as in the warm-up stage:
\begin{equation}
\mathcal{L}_{\mathrm{vq}}^{(2)}
=
\mathcal{L}_{\mathrm{vq}}^{(\mathrm{warm})}.
\end{equation}
The full objective becomes
\begin{equation}
\mathcal{L}^{(2)}
=
\mathcal{L}_{\mathrm{rec}}
+
\lambda_{\mathrm{perc}}\mathcal{L}_{\mathrm{perc}}
+
\lambda_{\mathrm{adv}}\mathcal{L}_{\mathrm{adv}}
+
\lambda_{\mathrm{vq}}\mathcal{L}_{\mathrm{vq}}^{(2)}.
\end{equation}
Because the encoder remains frozen, adversarial gradients can improve perceptual quality without reintroducing encoder drift. Stage~2 therefore separates perceptual refinement from latent-space movement, allowing the decoder and discriminator to improve visual fidelity while preserving the stabilized codebook geometry inherited from the warm-up stage.

Overall, NSVQ decomposes VQ training into three sequential phases:
\begin{equation*}
\text{collapse prevention}
\rightarrow
\text{frozen-encoder warm-up}
\rightarrow
\text{perceptual refinement}.
\end{equation*}
This staged design follows directly from the non-stationary view of collapse: first help the codebook follow the drifting encoder distribution, then freeze the encoder and warm up the quantizer--decoder pair under a stationary latent geometry, and finally reintroduce adversarial training for perceptual refinement.

\section{Experimental Setup}
\label{sec:experimental-setup}

We evaluate NSVQ on ImageNet-1k~\citep{russakovsky2015imagenet} reconstruction at $128\times128$ resolution. Following SimVQ~\citep{zhu2024addressing}, models are trained on the ImageNet training set and evaluated on the validation set without validation-based early stopping or checkpoint selection. Component-level ablations and hyperparameter studies are mainly conducted on CelebA~\citep{liu2015deep} at the same resolution due to computational constraints.

Our autoencoder follows the VQGAN-style encoder-decoder architecture~\citep{esser2021taming}. Unless otherwise specified, models use latent dimension $d=128$, codebook size $K=65{,}536$, and a $16\times16$ token grid. All compared ImageNet-1k autoencoder experiments use the same total training budget. We report rFID, LPIPS, SSIM, PSNR, and codebook utilization, where utilization is the fraction of codebook entries activated at least once during evaluation. Full details on tokenizer variants, optimization, stage transitions, and evaluation are provided in Appendix~\ref{app:experimental-details}.

\section{Results}
\label{sec:results}

We evaluate NSVQ on ImageNet-1k from three perspectives: reconstruction quality compared with representative VQ tokenizers, phase-level ablations of the proposed training strategy, and training dynamics that directly test the non-stationarity hypothesis. Additional CelebA ablations, hyperparameter sensitivity, detailed dynamics, and computational cost analyses are provided in Appendix~\ref{app:additional-results}.

\subsection{Main Results on ImageNet-1k}

Table~\ref{tab:main-results} compares NSVQ with representative VQ-based tokenizers on ImageNet-1k at $128\times128$ resolution. With the same codebook size of 65,536, NSVQ achieves the best overall reconstruction quality, obtaining 100.0\% codebook utilization, 2.10 rFID, 0.12 LPIPS, 25.21 PSNR, and 78.8\% SSIM. Compared with SimVQ \citep{zhu2024addressing}, NSVQ improves rFID from 2.39 to 2.10 and PSNR from 24.07 to 25.21 while maintaining full utilization. Since both SimVQ and NSVQ activate all codes, the improvement suggests that utilization alone is not sufficient to characterize codebook quality; stable alignment among the encoder distribution, codebook, and decoder is also important.

Applying the NS loss and the freeze-and-refine schedule to SimVQ improves its reconstruction quality, reducing rFID from 2.39 to 2.19 and improving LPIPS and PSNR, while maintaining high codebook utilization. This indicates that the proposed non-stationary-aware training strategy is not limited to our base VQ formulation, but can also benefit a strong SimVQ-style baseline. Nevertheless, this variant still underperforms full NSVQ, suggesting that the best performance is obtained when dense drift-aware updates, code replacement, encoder freezing, warm-up stabilization, and adversarial refinement are combined.

\begin{table}[t]
\centering
\caption{Reconstruction performance on ImageNet-1k at $128\times128$ resolution. All models encode images into $16\times16$ tokens. Codebook utilization is calculated as the fraction of codebook entries activated at least once on ImageNet validation. $^{\dagger}$FSQ uses a finite scalar quantization codebook with 64,000 effective codes rather than exactly 65,536 vector codes.}
\label{tab:main-results}
\begin{adjustbox}{width=\textwidth}
\begin{tabular}{lccccccc}
\toprule
Method & latent dim & tokens & Codebook size & Util.$\uparrow$ & rFID$\downarrow$ & LPIPS$\downarrow$ & PSNR$\uparrow$ / SSIM\%$\uparrow$ \\
\midrule
VQGAN \citep{esser2021taming} & 128 & 256 & 65,536 & 1.4\% & 3.74 & 0.17 & 22.20 / 70.6 \\
VQGAN-EMA \citep{razavi2019vqvae2} & 128 & 256 & 65,536 & 4.5\% & 3.23 & 0.15 & 22.89 / 72.3 \\
VQGAN-FC (d=128) \citep{yu2022vector} & 128 & 256 & 65,536 & 1.4\% & 5.33 & 0.18 & 21.45 / 68.8 \\
VQGAN-FC (d=8) \citep{yu2022vector} & 8 & 256 & 65,536 & 100.0\% & 2.63 & 0.13 & 23.79 / 77.5 \\
FSQ$^{\dagger}$ \citep{mentzer2024finite} & 6 & 256 & 64,000 & 100.0\% & 2.80 & 0.13 & 23.63 / 75.8 \\
LFQ \citep{yu2024language} & 16 & 256 & 65,536 & 100.0\% & 2.88 & 0.13 & 23.60 / 77.2 \\
\midrule
VQGAN-LC \citep{zhu2024scaling} & 128 & 256 & 65,536 & 100.0\% & 2.39 & 0.13 & 23.99 / 77.3 \\
SimVQ \citep{zhu2024addressing} & 128 & 256 & 65,536 & 100.0\% & 2.39 & 0.13 & 24.07 / 77.4 \\
FVQ \citep{chang2025scalable} & 128 & 256 & 65,536 & 100.0\% & 2.43 & 0.13 & 24.81 / 77.0 \\
SimVQ + NS loss + freeze-and-refine schedule & 128 & 256 & 65,536 & 99.2\% & 2.19 & 0.12 & 24.86 / 77.4 \\
\textbf{Ours} & 128 & \textbf{256} & \textbf{65,536} & \textbf{100.0\%} & \textbf{2.10} & \textbf{0.12} & \textbf{25.21 / 78.8} \\
\bottomrule
\end{tabular}
\end{adjustbox}
\end{table}
In addition to reconstruction at $128\times128$, we further evaluate whether the learned tokenizer benefits downstream latent diffusion generation and higher-resolution reconstruction. As shown in Table~\ref{tab:additional-imagenet-results}, under the same LDM U-Net trained for 100 epochs on ImageNet at $128\times128$, NSVQ achieves the best generation quality, reducing FID from 17.98 for VQGAN-LC and 20.11 for SimVQ to 16.62. On ImageNet $256\times256$ reconstruction, NSVQ also obtains the best overall reconstruction quality, improving rFID from 3.29--3.35 to 2.89 and achieving the best LPIPS, PSNR, and SSIM. These results indicate that the improved codebook stability of NSVQ transfers beyond the primary $128\times128$ reconstruction setting.

\begin{table}[t]
\centering
\caption{Additional ImageNet results. Left: downstream LDM generation on ImageNet \(128\times128\), using separately trained \(d=128\), \(K=65{,}536\) tokenizers with a \(16\times16\) token grid. The same latent-space U-Net is trained for 100 epochs for all tokenizers, and FID is computed from 5,000 generated samples. Right: ImageNet \(256\times256\) reconstruction, where all methods use downsampling factor 16 and therefore produce 256 tokens per image.}
\label{tab:additional-imagenet-results}
\small
\begin{adjustbox}{width=\textwidth}
\begin{tabular}{lcccc|lcccc}
\toprule
\multicolumn{5}{c|}{LDM generation at $128\times128$} &
\multicolumn{5}{c}{Reconstruction at $256\times256$} \\
\cmidrule(lr){1-5}\cmidrule(lr){6-10}
Method & tokens & Codebook size & Latent dim & FID$\downarrow$ &
Method & rFID$\downarrow$ & LPIPS$\downarrow$ & PSNR$\uparrow$ & SSIM\%$\uparrow$ \\
\midrule
VQGAN & 256 & 65,536 & 128 & 19.57 &
VQGAN-LC-ResNet & 3.30 & 0.13 & 22.50 & 57.48 \\
VQGAN-LC & 256 & 65,536 & 128 & 17.98 &
SimVQ & 3.29 & 0.13 & 22.58 & 57.56 \\
SimVQ & 256 & 65,536 & 128 & 20.11 &
FVQ & 3.35 & 0.13 & 23.28 & 57.26 \\
\textbf{Ours} & \textbf{256} & \textbf{65,536} & \textbf{128} & \textbf{16.62} &
\textbf{Ours} & \textbf{2.89} & \textbf{0.12} & \textbf{23.65} & \textbf{57.85} \\
\bottomrule
\end{tabular}
\end{adjustbox}
\end{table}

\subsection{Phase Ablation on ImageNet-1k}

Table~\ref{tab:imagenet-ablation} analyzes the role of each phase on ImageNet-1k. Stage~1 alone already achieves 100\% codebook utilization and competitive reconstruction quality, confirming that the non-stationary loss and code replacement mechanism prevent early collapse during joint encoder-quantizer training. Adding the frozen-encoder warm-up stage improves distortion-oriented metrics such as SSIM and PSNR, but worsens FID because adversarial training is disabled during this transition phase. This behavior is expected: the warm-up stage prioritizes quantizer--decoder alignment under a fixed encoder distribution rather than perceptual realism.

Directly skipping the warm-up stage and training with Stage~1 + Stage~2 degrades FID, SSIM, and PSNR compared with the full pipeline. This shows that adversarial refinement benefits from an intermediate collapse-safe transition after encoder freezing. The full NSVQ pipeline, consisting of Stage~1, warm-up, and Stage~2, achieves the best rFID while maintaining 100\% utilization.

Overall, these ablations show that the gain of NSVQ comes from matching each phase to the corresponding dynamics: drift-aware codebook learning, frozen-encoder stabilization, and adversarial perceptual refinement.

\begin{table}[t]
\centering
\caption{ImageNet ablation study of the proposed NSVQ training strategy. The full pipeline consists of Stage~1, a frozen-encoder warm-up stage, and Stage~2.}
\label{tab:imagenet-ablation}
\begin{adjustbox}{width=\textwidth}
\begin{tabular}{lcccccc}
\toprule
Variant & Training phases & rFID$\downarrow$ & LPIPS-VGG$\downarrow$ & SSIM$\uparrow$ & PSNR$\uparrow$ & Util.$\uparrow$ \\
\midrule
No NS loss, No Replacement & 1 only & 3.74  & 0.17 & 70.6  & 22.20 & 1.4\% \\
With NS loss, No Replacement & 1 only & 2.58  & 0.13 & 75.4  & 24.85 & 99.1\% \\
NSVQ Stage 1 only & 1 only & 2.24 & 0.12 & 78.8 & 25.15 & 100.0\% \\
NSVQ Stage 1 + Warm-up & 1 + W & 4.02 & \textbf{0.12} & \textbf{81.1} & \textbf{26.05} & 97.6\% \\
NSVQ Stage 1 + Stage 2 only & 1 + 2 & 2.59 & 0.13 & 75.4 & 24.86 & 99.1\% \\
Full NSVQ & 1 + W + 2 & \textbf{2.10} & \textbf{0.12} & 78.8 & 25.21 & \textbf{100.0\%} \\
\bottomrule
\end{tabular}
\end{adjustbox}
\end{table}
\subsection{Breaking the Drift--Collapse Feedback Loop}
\label{sec:drift-stress-test}

To directly test whether encoder drift can trigger the feedback loop described in
Section~\ref{sec:background_encoder_drift}, we conduct an encoder-drift stress test. We vary the
encoder learning-rate multiplier in
\(\{0.1,0.5,1,5,10\}\) while keeping the rest of the training configuration fixed. This intervention
provides a controlled way to increase encoder movement. We compare two settings: an unprotected
baseline without NS loss or codebook replacement, and the Stage-1 protection used by NSVQ, which
combines NS loss with codebook replacement.

We measure four quantities: the mean RMS encoder drift \(D_t^{\mathrm{RMS}}\), assignment churn,
final codebook utilization, and final rFID. Assignment churn measures how often latent positions
change their nearest-code assignments between consecutive logged checkpoints:
\begin{equation}
\mathrm{Churn}^{t}
=
\mathbb{E}_{x\in S,u}
\left[
\mathbf{1}
\left(
q_t(x,u) \neq q_{t-\Delta}(x,u)
\right)
\right],
\end{equation}
where \(S\) is a fixed evaluation subset and \(u\) indexes spatial latent positions.

\begin{figure}[t]
\centering
\includegraphics[width=\linewidth]{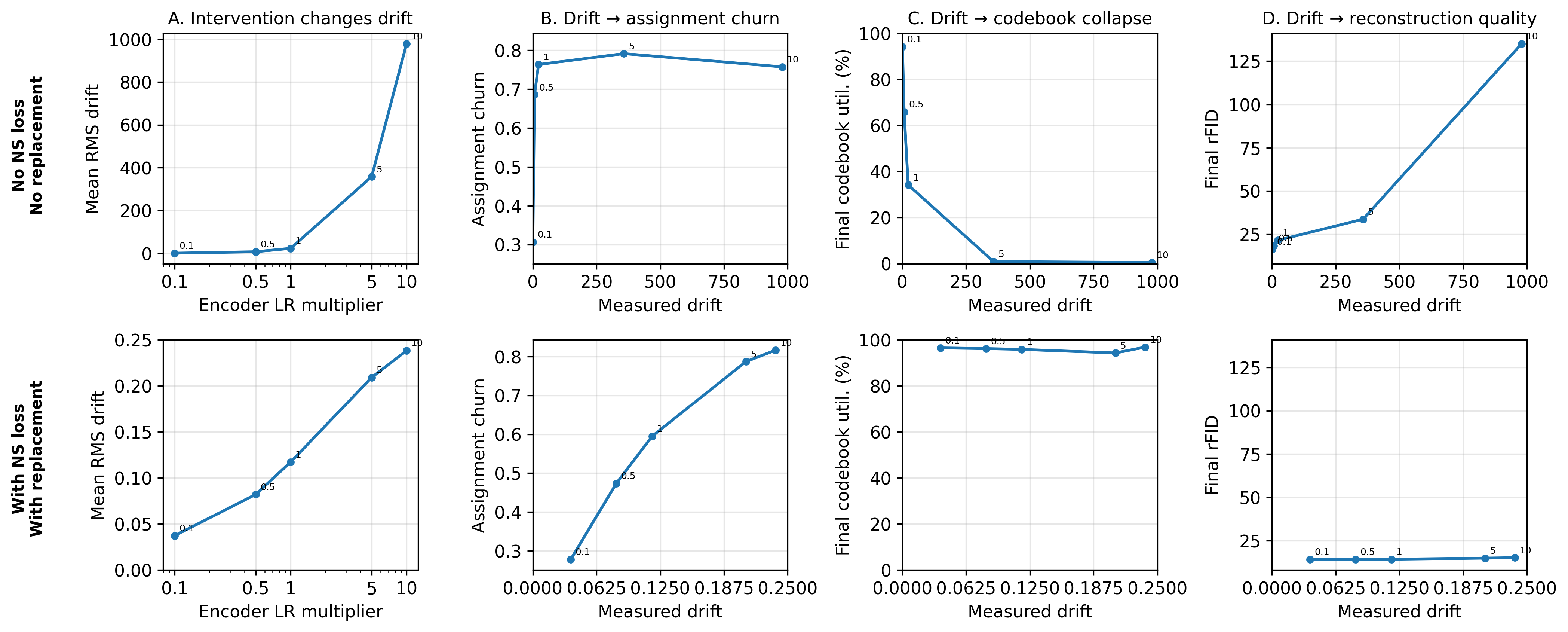}
\caption{Encoder-drift stress test. We vary the encoder learning-rate multiplier to intervene on
encoder drift. Top row: without NS loss and replacement, increasing encoder drift leads to high
assignment churn, codebook collapse, and catastrophic rFID degradation. Bottom row: with NS loss
and replacement, encoder drift and assignment churn still increase, but codebook utilization and rFID
remain stable. This indicates that NSVQ does not simply suppress encoder movement; instead, it
helps the codebook track the drifting latent distribution and prevents the drift--collapse feedback loop
from developing.}
\label{fig:drift-stress-test}
\end{figure}

Figure~\ref{fig:drift-stress-test} shows that, without NS loss and replacement, increasing the encoder
learning-rate multiplier substantially increases encoder drift. Assignment churn also increases, while
codebook utilization collapses and rFID deteriorates catastrophically. This behavior supports the
feedback-loop view: when the encoder moves faster than the sparsely updated codebook can follow,
codes lose assignments, the effective quantization quality degrades, and reconstruction quality can
collapse.

In contrast, with NS loss and replacement, encoder drift still increases as the encoder learning-rate
multiplier grows, and assignment churn also rises. However, codebook utilization and rFID remain
nearly stable. This shows that NSVQ does not simply suppress encoder movement. Instead, its Stage-1
protection helps the codebook track the drifting latent distribution, preventing assignment instability
from developing into codebook collapse and catastrophic reconstruction degradation. These results
provide direct evidence that compensating for encoder-induced non-stationarity can break the
drift--collapse feedback loop. Additional full-pipeline training curves showing the drop of measured
drift after encoder freezing are provided in Appendix~\ref{app:training-dynamics}.

\section{Discussion}
\label{sec:discussion}

Our results support the view that codebook collapse is a non-stationary alignment problem rather
than only a static utilization issue. During VQ training, the encoder distribution keeps moving, while
only selected code vectors receive direct updates. This mismatch can make inactive codes lag behind
and lose assignments. Once this occurs, collapse may become self-amplifying: codebook lag and dead
codes increase the quantization residual, which can enlarge the STE gradient-estimation gap and
further perturb encoder updates.

The encoder-LR stress test provides intervention evidence for this mechanism. Increasing the encoder
learning-rate multiplier increases drift; without NS loss and replacement, this leads to higher assignment
churn, severe codebook collapse, and catastrophic rFID degradation. Under NSVQ Stage-1 protection,
however, even substantially increased drift does not cause collapse or rFID degradation. Thus, drift
itself is not necessarily fatal; the failure arises when the codebook cannot track the moving latent
geometry and quantization error feeds back into encoder learning.

The results also show that high utilization alone is not sufficient. Several baselines and ablations use
nearly all codes, but still underperform the full NSVQ pipeline. Effective tokenizers therefore require
not only broad code usage, but also stable alignment among the encoder, codebook, and decoder.
NSVQ improves this alignment by matching each phase to the corresponding dynamics: drift-aware
codebook learning, frozen-encoder stabilization, and adversarial perceptual refinement.

Beyond reconstruction, NSVQ improves both $256\times256$ reconstruction and downstream latent
diffusion generation under the same U-Net training setup, suggesting that the learned discrete
representation benefits generative modeling as well. Limitations remain: our derivation uses a
latent-distance approximation to encoder-induced coupling; tighter theory and larger-scale
autoregressive, vision-language, and representation-learning evaluations are left for future work.

\section{Conclusion}

We present NSVQ, a non-stationary-aware training strategy for mitigating codebook collapse in VQ
models. Encoder drift can initiate a positive feedback loop: codebook lag produces dead codes and
larger quantization residuals, which amplify the STE gradient-estimation gap and further destabilize
encoder movement. NSVQ breaks this loop by helping the codebook track the moving latent geometry
before freezing the encoder for stable consolidation. Experiments on ImageNet-1k and CelebA show
that NSVQ achieves full codebook utilization and improves reconstruction quality over strong
baselines, suggesting that stable VQ training requires both broad code usage and robust
encoder--codebook--decoder alignment.

%% file: appendix.tex
\appendix
\section{Additional Analysis of Encoder-Induced Non-Stationarity}
\label{app:drift-analysis}

\subsection{Quantization Error and the STE Gradient-Estimation Gap}
\label{app:ste-gap}

The main text argues that codebook collapse can become self-amplifying because quantization error
feeds back into encoder updates through the straight-through estimator (STE). Here we provide a
more detailed derivation of this approximation.

Let the encoder output be $z_e=E_\theta(x)$ and the quantized representation be
\begin{equation}
z_q = z_e + \epsilon,
\qquad
\epsilon = z_q - z_e .
\end{equation}
Let $\ell(z)$ denote the decoder-side loss as a function of the latent input to the decoder. In a
continuous autoencoder, the encoder would receive a gradient evaluated at the continuous latent
point:
\begin{equation}
g_\theta^{\rm cont}
=
J_\theta(x)^\top \nabla_z \ell(z)\big|_{z=z_e},
\end{equation}
where
\begin{equation}
J_\theta(x)=\frac{\partial E_\theta(x)}{\partial \theta}
\end{equation}
is the encoder Jacobian. In VQ training with the STE, the forward pass uses $z_q$, but the quantizer
is treated as identity-like in the backward pass. Therefore, the encoder receives the decoder-side
gradient evaluated at the quantized point:
\begin{equation}
g_\theta^{\rm STE}
=
J_\theta(x)^\top \nabla_z \ell(z)\big|_{z=z_q}.
\end{equation}

Using a first-order Taylor expansion around $z_e$, we have
\begin{equation}
\nabla_z \ell(z_q)
=
\nabla_z \ell(z_e+\epsilon)
\approx
\nabla_z \ell(z_e) + H_\ell(z_e)\epsilon,
\end{equation}
where $H_\ell(z_e)$ is the Hessian of the decoder-side loss with respect to the latent input.
Substituting this approximation into the STE gradient gives
\begin{equation}
g_\theta^{\rm STE}
\approx
J_\theta(x)^\top \nabla_z \ell(z_e)
+
J_\theta(x)^\top H_\ell(z_e)\epsilon .
\end{equation}
Since the first term is $g_\theta^{\rm cont}$, the difference between the STE gradient and the
continuous-latent gradient is approximately
\begin{equation}
g_\theta^{\rm STE} - g_\theta^{\rm cont}
\approx
J_\theta(x)^\top H_\ell(z_e)\epsilon .
\end{equation}

This approximation shows that the STE gradient mismatch is controlled by the quantization residual
$\epsilon$ together with the local curvature of the decoder-side loss. When the codebook tracks the
encoder distribution well, $\|\epsilon\|$ remains small and the STE gradient is closer to the
continuous-latent gradient. However, when encoder drift causes codebook lag and dead codes,
the active code vectors become poorer representatives of the latent distribution, increasing
$\|\epsilon\|$. This can enlarge the STE gradient-estimation gap and further perturb encoder updates.

Thus, codebook collapse can form a positive feedback loop:
\begin{equation}
\text{encoder drift}
\rightarrow
\text{codebook lag}
\rightarrow
\text{dead codes}
\rightarrow
\|\epsilon\| \uparrow
\rightarrow
\text{STE gradient-estimation gap}
\rightarrow
\text{encoder drift} \uparrow .
\end{equation}
This provides the mechanism behind the catastrophic collapse behavior observed in the encoder-drift
stress test in Section~\ref{sec:drift-stress-test}.

\subsection{Batch-Size-One Illustration of Codebook Lag}
\label{app:batch-one-lag}

We also give a simple batch-size-one illustration of how encoder drift can make non-selected codes
lag behind the moving latent geometry. At step $t=0$, suppose the model processes sample $x_1$,
whose nearest code is $c_{q_1}^{(0)}$. The encoder is updated through the STE as
\begin{equation}
\theta^{(1)}
=
\theta^{(0)}
-
\eta_\theta
\nabla_\theta L(\theta)\big|_{\theta=\theta^{(0)}} .
\end{equation}
At the same time, the embedding loss updates only the winning code:
\begin{equation}
c_j^{(1)}
=
c_j^{(0)}
+
\eta_c
\left(z_e^{(0)}(x_1)-c_j^{(0)}\right)
\mathbf{1}[j=q_1].
\end{equation}
Therefore, after one training step, the encoder has changed globally, but only the selected code has
directly followed the movement of the latent space.

Now consider another sample $x_2$, whose nearest code at step $t=0$ is $c_{q_2}^{(0)}$, with
$q_2\neq q_1$. Because the encoder has already been updated using $x_1$, the latent position of
$x_2$ changes approximately as
\begin{equation}
z_e^{(1)}(x_2)
\approx
z_e^{(0)}(x_2)
+
J_\theta^{(0)}(x_2)\Delta\theta^{(0)} .
\end{equation}
However, because $c_{q_2}$ was not selected when processing $x_1$, it does not directly move:
\begin{equation}
c_{q_2}^{(1)} = c_{q_2}^{(0)} .
\end{equation}
This creates an immediate asymmetry: the latent feature of $x_2$ has drifted, but its previously
associated code has not. If the drift is large enough, $z_e^{(1)}(x_2)$ may cross a Voronoi boundary
and become closer to another code:
\begin{equation}
q^{(1)}(x_2)
=
\arg\min_j
\left\|
z_e^{(1)}(x_2)-c_j^{(1)}
\right\|_2^2
\neq
q_2 .
\end{equation}
Once this happens, $c_{q_2}$ loses the assignment. After a code stops winning assignments, it no
longer receives embedding updates and may fall increasingly far behind the moving latent distribution.
As discussed in Appendix~\ref{app:ste-gap}, such codebook lag can further increase the quantization
residual and feed back into encoder updates through the STE.

\subsection{Controlled Evidence Beyond Poor Initialization}
\label{app:controlled-drift}

To test whether utilization decay can occur even with favorable initialization, we train a continuous
autoencoder to convergence, extract its latent features, initialize the codebook by $k$-means, and then
continue VQ training with the STE. This setup removes poor codebook initialization as a primary
confounding factor: under a frozen encoder, the $k$-means centroids are already well aligned with the
latent distribution and form an approximate fixed point of the standard VQ update, as discussed in
Appendix~\ref{app:kmeans}.

Figure~\ref{fig:controlled-drift} shows that codebook utilization still decreases after VQ training begins,
while $D_{\rm RMS}^{t}$ remains non-zero. Thus, utilization decay is not merely a consequence of poor
initialization. Even from a favorable assignment structure, residual encoder movement after quantizer
insertion can destabilize the codebook. This observation is consistent with the feedback view in
Section~\ref{sec:background_encoder_drift}: encoder drift induces codebook lag, stale codes lose
assignments, quantization residuals grow, and the resulting STE gradient-estimation gap can further
destabilize the encoder.

\begin{figure}[t]
\centering
\includegraphics[width=0.72\linewidth]{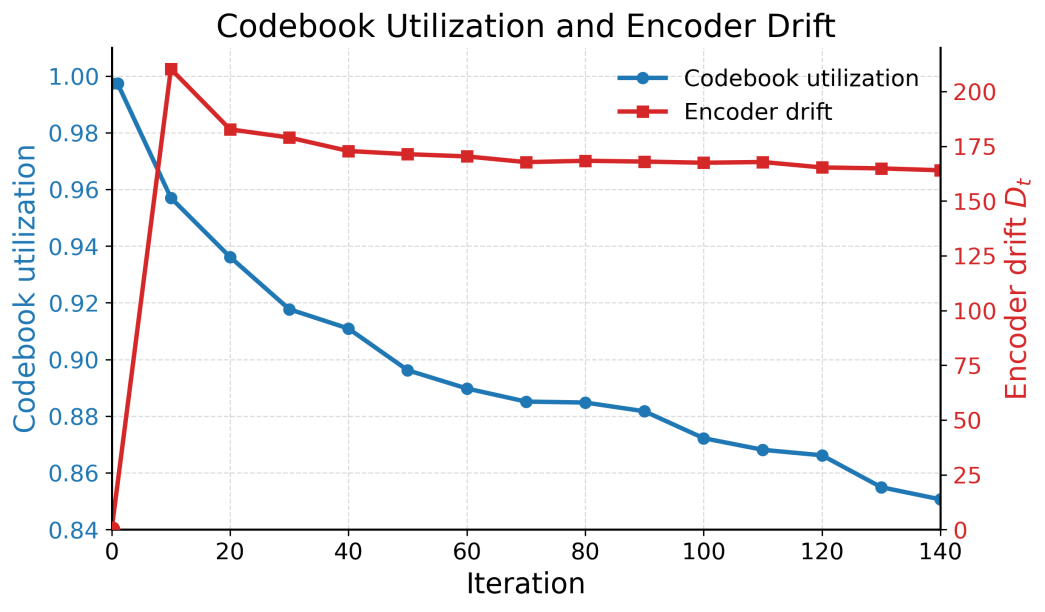}
\caption{Controlled evidence that encoder drift and codebook collapse remain coupled under favorable initialization. A continuous autoencoder is first trained to convergence, and the codebook is initialized by $k$-means on the converged latent features before STE-based VQ training begins. Despite this favorable initialization, the RMS encoder drift \(D_t^{\mathrm{RMS}}\) defined in Eq.~\eqref{eq:encoder-drift} remains non-zero and codebook utilization progressively decreases. Encoder drift is computed between consecutive logged checkpoints with interval \(\Delta=1\) optimizer steps on a fixed evaluation subset.}
\label{fig:controlled-drift}
\end{figure}

\section{Extended Related Work}
\label{app:extended-related-work}

\subsection{Stochastic Quantization, Regularization, and Code Maintenance}

A first line of work addresses codebook collapse by modifying the assignment or update rule so that more code vectors receive training signal. Instead of relying strictly on deterministic nearest-neighbor assignment, softened or stochastic quantization methods allow code selection to become probabilistic \citep{roy2018theory,williams2020hierarchical}. SQ-VAE \citep{takida2022sqvae}, for example, introduces stochastic quantization within a variational framework, allowing code selection to be treated probabilistically rather than as a hard argmin. Regularized VQ \citep{zhang2023regularized} further combines prior-distribution regularization and stochastic mask regularization to improve token utilization and reduce mismatch between training and inference.

Another pragmatic family of methods focuses on code maintenance. In large-scale systems, dead or stale codes are often reset or replaced using more active latent features. This idea is widely used in practice, including in Jukebox \citep{dhariwal2020jukebox} and SoundStream \citep{zeghidour2021soundstream}. Related work such as Online Clustered Codebook also updates the codebook with explicit clustering-style mechanisms \citep{zheng2023online}. Although such replacement heuristics do not by themselves explain the origin of collapse, they are effective at preventing permanent code starvation and are therefore a useful complement to more principled update mechanisms.

NSVQ adopts this practical viewpoint but places it in a non-stationary framework: replacement is not treated as a stand-alone solution, but rather as a way to compensate for the approximation error of the drift-propagation mechanism used during the early, highly non-stationary phase of training.

\subsection{Mapping-Based and Dense-Update Codebook Adaptation}

A second and increasingly influential line of work attempts to reduce collapse by making codebook updates denser. Instead of updating only the winning code vector, these methods parameterize the codebook through a transformation $P_{\psi}(\cdot)$, so that changing $\psi$ alters the entire set of effective codes simultaneously. Formally, let
\begin{equation}
C'=P_{\psi}(C).
\end{equation}
After observing one batch, not only the selected transformed code $P_{\psi}(c_{q_i})$ but all transformed codes can change because they share the same mapping parameters. In this sense, for any non-winning code $c_m$ with $m \ne q_i$, the update to $\psi$ induces an implicit motion term:
\begin{equation}
P_{\psi^{(t+1)}}(c_m)
=
P_{\psi^{(t)}}(c_m)
+
J_{\psi^{(t)}}(c_m)\Delta \psi^{(t)}.
\end{equation}

Straightening Out the Straight-Through Estimator \citep{huh2023straightening} is a key precursor in this direction. It introduces an affine reparameterization of code vectors to alleviate optimization difficulties caused by sparse codebook gradients and STE mismatch. VQGAN-LC \citep{zhu2024scaling} extends this idea to large codebooks by learning a projector over a frozen, pre-initialized codebook, enabling much higher utilization at scale. SimVQ \citep{zhu2024addressing} simplifies the mapping to a single linear layer and shows that even this minimal reparameterization can strongly improve codebook health. More recently, FVQ \citep{chang2025scalable} replaces the simple mapping with a more expressive projector architecture and reports full codebook utilization under large-scale training.

These methods are highly effective, but their dense updates are implicit. The induced motion of non-winning codes is not explicitly constrained to match the actual drift of the encoder distribution. In addition, mapping-based approaches introduce extra parameters and additional optimization dynamics, which may increase computational cost or create new capacity bottlenecks. NSVQ is complementary to this family: rather than relying only on a learned projector, it explicitly models the non-stationary movement of the encoder distribution and encourages nearby code vectors to track it during the unstable early phase of training.

\subsection{Freezing the Encoder and Continuing to Optimize the Quantizer or Decoder}

A third direction decouples representation learning from discretization by freezing some or all of the backbone once the continuous latent space is sufficiently mature. This idea aligns closely with our non-stationarity argument: if encoder drift is a major driver of collapse, then stopping encoder updates after convergence should make codebook learning substantially easier.

Recent tokenizer research increasingly follows this philosophy. CODA \citep{liu2025coda} studies how to convert off-the-shelf continuous VAEs into discrete tokenizers, and its ablations show that adapting the decoder yields substantial gains, whereas adapting the encoder provides only marginal benefit. ReVQ \citep{zhang2025revq} freezes both the encoder and decoder of a pretrained VAE and trains only a quantizer together with a lightweight rectifier. VQ-Transplant \citep{fang2026vqtransplant} preserves pretrained encoder-decoder parameters, replaces the quantization module, and introduces a short decoder adaptation stage to reduce the mismatch between the new quantizer and the pretrained decoder. These works suggest that, once the latent space is sufficiently mature, continued encoder updates may no longer be essential for improving discretization quality. Instead, fixing the latent geometry and optimizing the quantizer and decoder around it can lead to a favorable stability-performance trade-off.

QLIP \citep{zhao2025qlip} also adopts a two-stage training pipeline in which the visual encoder is frozen in the second stage and only the bottleneck quantizer and decoder are optimized. However, its motivation differs from ours. QLIP freezes the encoder primarily to resolve practical conflicts between contrastive learning and GAN-based reconstruction, which prefer different batch-size and memory regimes. In contrast, our freezing strategy is motivated by the dynamics of VQ training itself. We freeze the encoder to remove encoder-induced latent drift, thereby converting quantizer learning into a more stable problem under a fixed latent geometry.

Although freezing the encoder can stabilize tokenizer training, prior work also suggests that naive frozen-backbone quantization is not always sufficient for high-quality VQ reconstruction \citep{liu2025coda,zhang2025revq,fang2026vqtransplant}. In standard VQ-VAE training, the commitment loss shapes encoder outputs toward the selected code vectors \citep{oord2017neural}. If the encoder is frozen from the beginning, this objective cannot adapt the continuous feature space to the discrete codebook. As a result, the quantizer must fit a fixed latent distribution that may not be sufficiently clustered, fine-grained, or reconstruction-oriented. Consistent with this limitation, recent studies on continuous-to-discrete tokenizer conversion show that additional quantizer, decoder, or rectifier adaptation is often needed to recover fine-grained reconstruction quality \citep{liu2025coda,zhang2025revq,wang2025tokenbridge}.

Our stage-wise freezing strategy differs from naive frozen-backbone tokenization. NSVQ does not freeze an arbitrary pretrained encoder at the start. Instead, Stage~1 first trains the encoder, decoder, and quantizer jointly under the VQ objective, allowing the latent space to become reconstruction-oriented and quantization-aware. This design is also consistent with recent evidence that the timing of quantization matters: allowing the encoder to first form a more mature latent space can improve the quality of the discrete codebook \citep{liu2025coda,zhang2025revq,wang2025tokenbridge}. Only after the encoder has largely matured do we freeze it after Stage~1. At this point, freezing no longer prevents the encoder from learning a VQ-compatible representation; rather, it removes further encoder drift and converts codebook learning into a stable consolidation problem. Stage~2 then reintroduces adversarial refinement under the same fixed latent geometry. Therefore, in NSVQ, encoder freezing is not used as a stand-alone shortcut, but as a carefully timed mechanism for suppressing non-stationarity after the latent space has already been shaped by VQ training.

\section{Additional Details of the NSVQ Training Strategy}
\label{app:nsvq-details}

\subsection{VQ Autoencoder Setup}

NSVQ builds on a standard VQ-based autoencoder with an encoder $E_{\theta}$, a decoder $D_{\phi}$, and a discrete codebook $C=\{c_j\}_{j=1}^{K}$, where $c_j\in\mathbb{R}^{d}$. Given an input image $x$, the encoder produces a continuous latent representation
\begin{equation}
z_e=E_{\theta}(x).
\end{equation}
Quantization is performed by nearest-neighbor assignment:
\begin{equation}
q=\arg\min_j \|z_e-c_j\|_2^2,
\end{equation}
and the decoder reconstructs the image from the quantized representation $z_q=c_q$:
\begin{equation}
\hat{x}=D_{\phi}(z_q).
\end{equation}
\subsection{Derivation of the Non-Stationary Embedding Loss}
\label{app:nsvq-derivation}

The core idea of NSVQ is to propagate encoder drift not only to the winning code, but also to nearby non-winning codes, so that the codebook can better track a non-stationary latent distribution. In standard VQ training, only the selected code receives a direct embedding update for each latent vector. However, when the encoder changes rapidly, nearby codes may also need to move with the local latent manifold; otherwise, they can lag behind the drifting encoder distribution and eventually become inactive.

At iteration $t$, suppose the current sample is $x_i$ with encoder output $E_{\theta^{(t)}}(x_i)$, and let $q_i$ be its winning code index. For another sample $x_j$, the encoder update induced by $x_i$ changes its latent representation as
\begin{equation}
E_{\theta^{(t+1)}}(x_j)
=
E_{\theta^{(t)}}(x_j)
+
\Delta E(x_j).
\end{equation}
Under a first-order approximation,
\begin{equation}
\Delta E(x_j)
\approx
J_{\theta}^{(t)}(x_j)\Delta\theta^{(t)},
\end{equation}
where $J_{\theta}^{(t)}(x_j)$ is the encoder Jacobian with respect to the parameters. One gradient step induced by the loss on $x_i$ gives
\begin{equation}
\Delta\theta^{(t)}
=
-\eta
\frac{\partial \mathcal{L}(E_{\theta^{(t)}}(x_i))}{\partial \theta}
=
-\eta
J_{\theta}^{(t)}(x_i)^\top g_i,
\qquad
g_i
=
\frac{\partial \mathcal{L}}{\partial E_{\theta^{(t)}}(x_i)}.
\end{equation}
Substituting this into the expression for $\Delta E(x_j)$ yields
\begin{equation}
\Delta E(x_j)
\approx
-\eta
J_{\theta}^{(t)}(x_j)
J_{\theta}^{(t)}(x_i)^\top
g_i
=
-\eta
K_{\theta}^{(t)}(x_j,x_i)g_i,
\end{equation}
where
\begin{equation}
K_{\theta}^{(t)}(x_j,x_i)
=
J_{\theta}^{(t)}(x_j)
J_{\theta}^{(t)}(x_i)^\top
\end{equation}
is the encoder-induced neural tangent kernel between $x_j$ and $x_i$. This expression shows that the latent movement of $x_j$ caused by an update on $x_i$ depends on the coupling between their encoder Jacobians.

Computing this exact kernel is intractable during large-scale VQ training. We therefore approximate the coupling between latent points using a distance-based kernel in latent space:
\begin{equation}
k(x_j,x_i)
=
\exp
\left(
-\frac{
\|E_{\theta}(x_j)-E_{\theta}(x_i)\|_2^2
}{
2\sigma^2
}
\right).
\label{eq:kernel}
\end{equation}
This replacement approximates the matrix-valued NTK \(K_\theta(x_j,x_i)\) by an isotropic scalar kernel \(k(x_j,x_i)I\). We use this only as a computationally tractable locality approximation, not as an exact NTK estimate. This gives the practical approximation
\begin{equation}
\Delta E(x_j) \approx k(x_j,x_i)\Delta E(x_i).
\label{eq:drift_estimation}
\end{equation}
Thus, Eq.~\eqref{eq:drift_estimation} motivates where to propagate auxiliary training signal:
latent locations close to the current sample should receive stronger drift propagation than distant
locations. However, the NTK-style analysis does not directly specify the codebook update direction,
because the exact task-gradient-induced drift vector \(\Delta E(x_i)\) is not explicitly available for
each non-winning code.

We therefore use a practical drift-and-catch-up surrogate for the update direction. The residual
\(\operatorname{sg}[z_i]-c_{q_i}\) approximates the local encoder-codebook mismatch around the
currently selected active code \(c_{q_i}\). However, directly translating a non-winning code by this
residual implicitly assumes that the code is already active and associated with a nearby latent point.
This assumption is undesirable in large-codebook training, where some codes may be stale or inactive.
To prevent such codes from falling further behind, we add a catch-up correction \(c_{q_i}-c_j\), which
pulls the non-winning code toward the selected active code. Combining the two terms gives
\[
(\operatorname{sg}[z_i]-c_{q_i}) + (c_{q_i}-c_j)
=
\operatorname{sg}[z_i]-c_j .
\]
Therefore, the final auxiliary direction is a locality-weighted drift-and-catch-up update:
\begin{equation}
\Delta c_j
\propto
w_{ij}
\left(\operatorname{sg}[z_i]-c_j\right),
\qquad
j\ne q_i,
\end{equation}
where
\begin{equation}
w_{ij}
=
\exp\left(
-\frac{\|\operatorname{sg}[z_i]-c_j\|_2^2}{2\sigma^2}
\right).
\end{equation}
The winning code is still updated by the standard embedding loss, while nearby non-winning codes receive auxiliary drift-aware updates. This yields the following update rule:
\begin{equation}
\Delta c_j
=
\eta_c
\begin{cases}
\operatorname{sg}[z_i]-c_{q_i}, & j=q_i,\\[3pt]
w_{ij}\left(\operatorname{sg}[z_i]-c_j\right), & j\ne q_i,
\end{cases}
\label{eq:nsvq-codebook-update}
\end{equation}
where \(w_{ij}\) is the detached locality weight.
The non-winning branch of Eq.~\eqref{eq:nsvq-codebook-update} can be implemented by minimizing the following auxiliary embedding objective with detached locality weights:
\begin{equation}
\mathcal{L}'_{\text{emb}}
=
\|\operatorname{sg}[E_\theta(x_i)]-c_{q_i}\|_2^2
+
\sum_{j\ne q_i}
\operatorname{sg}\!\left[
\exp\left(
-\frac{
\|\operatorname{sg}[E_\theta(x_i)]-c_j\|_2^2
}{2\sigma^2}
\right)
\right]
\|\operatorname{sg}[E_\theta(x_i)]-c_j\|_2^2 .
\label{eq:aux-emb-stopgrad}
\end{equation}
Because \(w_{ij}\) is detached, the gradient with respect to \(c_j\) is proportional to
\(w_{ij}(c_j-\operatorname{sg}[z_i])\), so gradient descent produces the attractive update
\(w_{ij}(\operatorname{sg}[z_i]-c_j)\). The weight determines which non-winning codes are close enough to receive an auxiliary update, but it is not differentiated. Therefore, for \(j\ne q_i\), the negative gradient direction is
\[
-\nabla_{c_j}\mathcal{L}'_{\text{emb}}
\propto
w_{ij}\left(\operatorname{sg}[E_\theta(x_i)]-c_j\right),
\]
where \(w_{ij}\) denotes the detached locality weight. Hence the induced update is strictly attractive toward the current encoder output.
The first term is the standard embedding loss for the winning code. The second term extends the update to nearby non-winning codes, with a strength determined by their distance to the current encoder output.

In the implementation used in the main paper, we further normalize the kernel weights for easier tuning. For each latent vector $z_i=E_{\theta}(x_i)$, we define
\begin{equation}
d_{ij}
=
\left\|
\operatorname{sg}[z_i]-c_j
\right\|_2^2,
\qquad
q_i=\arg\min_j d_{ij},
\end{equation}
and use the softmax-normalized locality weight
\begin{equation}
p_{ij}
=
\frac{\exp(-d_{ij}/\tau)}
{\sum_{k=1}^{K}\exp(-d_{ik}/\tau)},
\qquad
\tilde p_{ij}=\operatorname{sg}[p_{ij}] .
\label{eq:softmax-locality-stopgrad}
\end{equation}
Here, $\tau$ plays the role of a temperature parameter that controls the locality of the dense update. A smaller $\tau$ concentrates the auxiliary update on codes closest to the current latent vector, while a larger $\tau$ spreads the update more broadly. As in the main text, the softmax is normalized over all codes, including the winning code, while the auxiliary loss sums only over non-winning codes. We do not redistribute the winning-code probability mass, because the winning code already receives the standard embedding update.

The practical non-stationary embedding loss used by NSVQ is
\begin{equation}
\mathcal{L}_{NS}
=
\frac{1}{Nd}
\sum_{i=1}^{N}
\sum_{j\ne q_i}
\tilde p_{ij}d_{ij}.
\label{eq:practical-ns-loss}
\end{equation}
where $N$ is the number of latent positions and $d$ is the code dimension. This loss corresponds to the softmax-normalized version of the kernel-weighted non-winning-code update, with the normalized locality weights treated as constants during backpropagation. The stop-gradient prevents the derivative of the softmax weights from introducing repulsive terms and makes the induced codebook update consistent with the attractive update rule in Eq.~\eqref{eq:nsvq-codebook-update}. The normalization makes the dense-update term easier to tune, because its global contribution can be controlled by a single coefficient $\alpha$.

Combining the standard winning-code embedding loss, the commitment loss, and the non-stationary dense update gives the complete Stage-1 quantization loss:
\begin{equation}
\mathcal{L}_{\mathrm{vq}}^{(1)}
=
\frac{1}{Nd}
\sum_{i=1}^{N}
\left\|
\operatorname{sg}[z_i]-c_{q_i}
\right\|_2^2
+
\beta
\frac{1}{Nd}
\sum_{i=1}^{N}
\left\|
z_i-\operatorname{sg}[c_{q_i}]
\right\|_2^2
+
\alpha
\mathcal{L}_{\mathrm{NS}}.
\end{equation}
Equivalently, using $z_q$ to denote the quantized latent representation, this can be written compactly as
\begin{equation}
\mathcal{L}_{\mathrm{vq}}^{(1)}
=
\left\|
z_q-\operatorname{sg}[z_e]
\right\|_2^2
+
\beta
\left\|
z_e-\operatorname{sg}[z_q]
\right\|_2^2
+
\alpha
\mathcal{L}_{\mathrm{NS}}.
\end{equation}
Thus, NSVQ preserves hard nearest-neighbor quantization in the forward pass, while adding a drift-aware auxiliary update for nearby non-winning codes during the non-stationary Stage~1 training phase.

\subsection{Codebook Replacement in Stage 1}

In addition to the non-stationary embedding loss, Stage~1 uses a codebook replacement strategy to revive persistently inactive entries. During each epoch, we accumulate the usage count of every code. At the end of the epoch, codes whose usage falls below a dead-code threshold are marked as inactive. Each dead code is replaced by copying an active source code, optionally with a small Gaussian perturbation:
\begin{equation}
c_j^{\mathrm{new}}
=
c_{s(j)}
+
\epsilon,
\qquad
\epsilon\sim\mathcal{N}(0,\sigma^2I).
\end{equation}
The source code $c_{s(j)}$ can be selected either from the most frequently used code or by sampling from active codes with probability proportional to their usage. In our implementation, replacement is used only in Stage~1, where encoder drift is strongest and dead-code recovery is most necessary. After the encoder is frozen, replacement is disabled because the codebook is then optimized under a fixed latent geometry.

\subsection{Encoder-Freezing Criterion}

We use the epoch-averaged commitment loss as a practical indicator of whether the encoder--codebook interaction has stabilized. Let $\ell_t^{\mathrm{commit}}$ denote the average commitment loss over epoch $t$:
\begin{equation}
\ell_t^{\mathrm{commit}}
=
\frac{1}{|\mathcal{B}_t|}
\sum_{b\in\mathcal{B}_t}
\mathcal{L}_{\mathrm{commit}}(b),
\end{equation}
where $\mathcal{B}_t$ is the set of mini-batches in epoch $t$, and
\begin{equation}
\mathcal{L}_{\mathrm{commit}}
=
\left\|
z_e-\operatorname{sg}[z_q]
\right\|_2^2.
\end{equation}
The encoder is frozen when the epoch-averaged commitment loss has not decreased for 10 consecutive epochs. Equivalently, if no new lower value of $\ell_t^{\mathrm{commit}}$ is observed within a 10-epoch patience window, we regard the encoder--codebook interaction as having reached a plateau:
\begin{equation}
\min_{s=t-9}^{t}
\ell_s^{\mathrm{commit}}
\ge
\min_{u\le t-10}
\ell_u^{\mathrm{commit}}.
\end{equation}
A decreasing commitment loss indicates that the encoder outputs and selected code vectors are still adapting to each other. In contrast, when the epoch-averaged commitment loss does not decrease for 10 consecutive epochs, the latent geometry is considered sufficiently stable. NSVQ then freezes the encoder and enters the frozen-encoder warm-up stage.

\subsection{Why the Warm-up Stage Is Separated from Stage 2}

A direct transition from Stage~1 to Stage~2 can be unstable because adversarial gradients may disrupt codebook and decoder adaptation immediately after the encoder is frozen. Although the latent distribution becomes stationary after freezing, the quantizer and decoder may still be insufficiently aligned with this fixed encoder manifold. Applying GAN pressure too early can amplify transient quantization errors and destabilize reconstruction.

The warm-up stage avoids this issue by first optimizing the quantizer and decoder without adversarial interference. During warm-up, the encoder remains frozen, while adversarial training, $\mathcal{L}_{\mathrm{NS}}$, and codebook replacement are disabled. Once the quantizer and decoder have adapted to the fixed latent geometry, Stage~2 re-enables adversarial training for perceptual refinement. Thus, the warm-up stage acts as a collapse-safe transition between non-stationary codebook learning and adversarial refinement.

\subsection{Implementation Details for Stage Transitions}

To ensure stable and reproducible training, we carefully control trainable parameters and adversarial updates during stage transitions. When transitioning from Stage~1 to the warm-up stage, the encoder parameters are frozen by setting \texttt{requires\_grad=False}. The decoder and quantizer remain trainable. The non-stationary embedding loss and codebook replacement are disabled, since encoder-induced non-stationarity has been removed. Adversarial training is also disabled entirely during warm-up, so both discriminator updates and generator-side adversarial losses are removed.

When transitioning from the warm-up stage to Stage~2, the encoder remains frozen and adversarial training is re-enabled. In practice, we use a short linear ramp for the adversarial loss so that the generator and discriminator are reintroduced gradually rather than abruptly. Learning rates and optimizer hyperparameters are kept unchanged across phases. These choices ensure that phase transitions modify only the intended aspects of training, such as removing encoder drift or reintroducing perceptual refinement, rather than introducing unrelated optimization artifacts.
\section{Additional Experimental Details}
\label{app:experimental-details}

\subsection{Datasets and Evaluation Protocol}

We evaluate NSVQ on image reconstruction using ImageNet-1k at $128\times128$ resolution. Following SimVQ \citep{zhu2024addressing}, models are trained on the ImageNet training set and evaluated on the ImageNet validation set. We do not use validation-based early stopping or checkpoint selection. Due to computational constraints, controlled component-level ablations and hyperparameter studies are mainly conducted on CelebA at $128\times128$ resolution, while the main comparison is reported on ImageNet-1k.

For fair internal comparisons and ablations, we keep the training configuration fixed unless a specific component is being evaluated. Codebook utilization is computed as the fraction of codebook entries activated at least once on the evaluation set.

\subsection{Architecture and Quantizer Hyperparameters}

Our VQ autoencoder follows the VQGAN-style encoder-decoder architecture \citep{esser2021taming}. Unless otherwise specified, the latent channel dimension is $d=128$ and the codebook size is $K=65{,}536$. All models encode images into $16\times16$ discrete tokens. The encoder-decoder backbone uses a base channel width of 128 and two residual blocks at each resolution level.

The NSVQ quantizer uses commitment coefficient $\beta=0.25$, non-stationary loss coefficient $\alpha=0.1$, and temperature $\tau=0.35$. Unless otherwise specified, the loss weights are set to
\begin{equation}
\lambda_{\mathrm{perc}}=\lambda_{\mathrm{adv}}=\lambda_{\mathrm{vq}}=1.
\end{equation}

\paragraph{Additional tokenizer configurations.}
Unless otherwise specified, the default tokenizer uses latent channel dimension \(d=128\), codebook size \(K=65{,}536\), and a \(16\times16\) token grid. For the LDM generation experiment, we train a separate tokenizer with latent channel dimension \(d=128\), codebook size \(K=65{,}536\), and a \(16\times16\) token grid on ImageNet-1k at \(128\times128\) resolution. 

For the \(256\times256\) reconstruction experiment, we train separate tokenizers at \(256\times256\) resolution with a spatial downsampling factor of \(16\). This produces a \(16\times16\) token grid, corresponding to 256 discrete tokens per image. All compared methods in this setting use the same resolution, downsampling factor, token budget, codebook size, and evaluation protocol.

\subsection{Stage 1 + Warm-up + Stage 2 Schedule}

NSVQ is trained with the proposed Stage~1 + Warm-up + Stage~2 schedule. All ImageNet-1k \(128\times128\) autoencoder experiments are trained for \(70\) epochs, corresponding to \(700700\) optimizer steps under the global batch size specified in Appendix~D.4. CelebA \(128\times128\) experiments are trained for \(100\) epochs, corresponding to \(90700\) optimizer steps. The same total budget is used for all methods within each dataset unless otherwise specified. Stage~1 corresponds to non-stationary-aware codebook learning. During this phase, the encoder, decoder, quantizer, and discriminator are jointly trained. The non-stationary embedding loss and codebook replacement are enabled to prevent early codebook collapse while the encoder distribution is still drifting.

Codebook replacement is performed only during Stage~1. We accumulate the usage count of each code during every epoch. After a 5-epoch replacement warm-up, codes with usage below 1 are treated as inactive and replaced by active codes using probabilistic sampling with Gaussian perturbation of standard deviation 0.001. Replacement is disabled after encoder freezing, because the latent distribution is then fixed and the codebook can be optimized by standard fitting losses.

The transition from Stage~1 to the frozen-encoder warm-up stage is determined by a commitment-loss plateau criterion. Let $\ell_t^{\mathrm{commit}}$ denote the average commitment loss over epoch $t$. We freeze the encoder when $\ell_t^{\mathrm{commit}}$ has not decreased for 10 consecutive epochs. Equivalently, if no new lower value of $\ell_t^{\mathrm{commit}}$ is observed within a 10-epoch patience window, we regard the encoder--codebook interaction as stabilized and freeze the encoder. In our ImageNet experiments, this criterion triggers encoder freezing around epoch~50. The detailed step-based training-dynamics figures in Appendix~E.3 are reported on CelebA
\(128\times128\), where the same criterion triggers at step~45{,}350.

After encoder freezing, NSVQ enters a short frozen-encoder warm-up stage for 3 epochs. During warm-up, the encoder remains frozen, while the decoder and quantizer remain trainable. Adversarial training, $\mathcal{L}_{\mathrm{NS}}$, and codebook replacement are disabled. This phase allows the quantizer and decoder to adapt to the fixed encoder manifold using reconstruction-oriented losses before GAN pressure is reintroduced.

Stage~2 continues training with the encoder frozen. The decoder and quantizer remain trainable, and adversarial training is re-enabled for perceptual refinement. Given the fixed total budget of \(70\) epochs, Stage~2 is run for the remaining epochs after the commitment-loss-triggered Stage~1 and the 3-epoch warm-up stage. In our ImageNet experiments, the freeze criterion typically triggers around epoch~50, so Stage~2 lasts approximately \(17\) epochs. To avoid abrupt optimization changes, the adversarial loss is linearly ramped up over the first 2 epochs of Stage~2, with a maximum GAN factor of 1.0. The discriminator is not reset when transitioning from warm-up to Stage~2.

\subsection{Optimization Details}

All models are trained with Adam using a learning rate of $1\times10^{-4}$ and no learning-rate scheduler. Exponential moving average (EMA) is enabled throughout training. Training is conducted with distributed data parallelism on 4 GPUs using mixed precision. The batch size is 32 per GPU, yielding a global batch size of 128. The number of dataloader workers is 4 per process.

Across phase transitions, we preserve optimizer states for the remaining trainable parameters. When the encoder is frozen, its parameters are removed from gradient updates by setting \texttt{requires\_grad=False}. The decoder and quantizer remain trainable throughout all phases. During the warm-up stage, discriminator updates and generator-side adversarial losses are disabled entirely. During Stage~2, adversarial training is re-enabled while keeping the encoder frozen.

\subsection{LDM Generation Setup}
\paragraph{LDM generation setup.}
For the LDM generation experiment, we train a separate tokenizer on ImageNet-1k at
\(128\times128\) resolution and use it as the first-stage model for latent diffusion training.
The latent diffusion model operates on a \(16\times16\) latent grid with 128 latent channels.
For all tokenizers, we use the same latent-space U-Net architecture: input and output channels
are both 128, the base channel width is 256, the channel multipliers are \([1,2,4]\), each
resolution level uses 2 residual blocks, attention is applied at latent resolutions \(16,8,4\),
the number of attention heads is 4, and dropout is set to 0.0.

The diffusion model is trained with a DDPM objective using 1000 diffusion timesteps, a linear
noise schedule from \(1\times10^{-4}\) to \(2\times10^{-2}\), \(\epsilon\)-prediction, and an
\(L_2\) denoising loss. We train the LDM for 100 epochs using AdamW with learning rate
\(1\times10^{-4}\), weight decay 0.0, mixed precision training, and distributed data parallelism
on 4 GPUs. The data-loader batch size is 32 per GPU, giving a global batch size of 128. During
evaluation, samples are generated using deterministic DDIM sampling with 50 steps
(\(\eta=0.0\)). FID is computed using 5,000 generated samples against the ImageNet validation
set. Here, \(\eta=0.0\) denotes deterministic DDIM sampling without additional stochastic noise
in the reverse process.

\subsection{Evaluation Metrics}

We evaluate reconstruction quality using rFID, LPIPS, SSIM, and PSNR. Codebook utilization is computed as the fraction of codebook entries activated at least once on the evaluation set. We also report computational cost, including average epoch time and peak GPU memory usage, measured under the same distributed training setup.

\section{Additional Results}
\label{app:additional-results}

\subsection{Component-Level Ablation on CelebA}
\label{app:celeba-ablation}

To analyze the contribution of each component under limited compute, we conduct a controlled ablation study on CelebA at $128\times128$ resolution. Table~\ref{tab:celeba-ablation} shows that no single component is sufficient to achieve the best performance. The freeze-and-refine schedule alone leaves codebook utilization low, indicating that encoder freezing cannot recover a poorly trained codebook by itself. Code replacement improves utilization, while the non-stationary loss achieves better FID, suggesting that dense drift-aware updates more directly address the mismatch caused by encoder drift. The full NSVQ pipeline achieves the best FID and LPIPS while maintaining high utilization, showing that the components are complementary.

\begin{table}[h]
\centering
\caption{Ablation study of NSVQ on CelebA at $128\times128$ resolution. The full pipeline consists of Stage~1, a frozen-encoder warm-up stage, and Stage~2. Lower is better for FID and LPIPS; higher is better for SSIM, PSNR, and utilization.}
\label{tab:celeba-ablation}
\begin{adjustbox}{width=\textwidth}
\begin{tabular}{lccccccccc}
\toprule
Variant & Freeze-and-refine schedule & NS loss & Codebook replacement & Training phases & FID$\downarrow$ & LPIPS$\downarrow$ & SSIM$\uparrow$ & PSNR$\uparrow$ & Util.$\uparrow$ \\
\midrule
Freeze schedule only & \checkmark & -- & -- & 1, W, 2 & 14.87 & 0.095 & 85.5 & 27.14 & 0.36 \\
Replacement only & -- & -- & \checkmark & 1 only & 15.38  & 0.089 & 86.6 & 27.35 & 0.82   \\
NS loss only & -- & \checkmark & -- & 1 only & 14.43 & 0.084 & 86.4 & 27.99 & 0.84 \\
Stage 1 + Stage 2 only & \checkmark & \checkmark & \checkmark & 1 + 2 & 14.20 & 0.085 & 87.1 & 27.60 & 0.88 \\
Stage 1 + Warm-up only & \checkmark & \checkmark & \checkmark & 1 + W & 17.26 & \textbf{0.083} & \textbf{88.3} & \textbf{29.00} & 0.93 \\
No encoder freezing & -- & \checkmark & \checkmark & 1 only & 14.37 & 0.088 & 86.8 & 28.24 & \textbf{0.99} \\
\textbf{Full NSVQ} & \checkmark & \checkmark & \checkmark & 1, W, 2 & \textbf{13.42} & \textbf{0.079} & 87.9 & 27.98 & \textbf{0.99} \\
\bottomrule
\end{tabular}
\end{adjustbox}
\end{table}

\subsection{Hyperparameter Sensitivity}
\label{app:sensitivity}

\begin{figure}[h]
\centering
\includegraphics[width=0.75\linewidth]{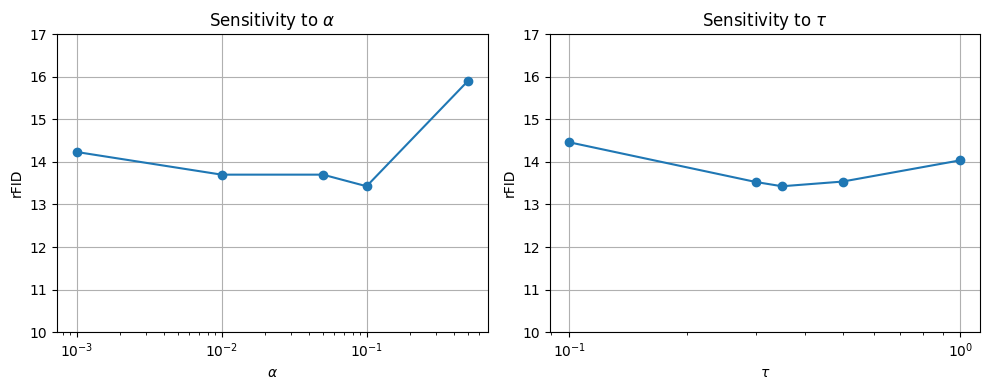}
\caption{Sensitivity analysis of NSVQ hyperparameters.}
\label{fig:sensitivity}
\end{figure}

We evaluate the sensitivity of NSVQ to the dense-update coefficient $\alpha$ and temperature $\tau$ in the non-stationary loss. As shown in Figure~\ref{fig:sensitivity}, NSVQ is stable across a moderate range of $\alpha$, with $\alpha=0.01$, $0.05$, and $0.1$ all achieving competitive rFID. The best result is obtained at $\alpha=0.1$. For $\tau$, both $\tau=0.3$ and $\tau=0.35$ perform well, while overly small or large temperatures degrade performance. A very small $\tau$ makes the dense update too local, whereas a very large $\tau$ spreads the update too broadly and weakens locality.

\subsection{Detailed Training Dynamics}
\label{app:training-dynamics}

\begin{figure}[h]
\centering
\includegraphics[width=\linewidth]{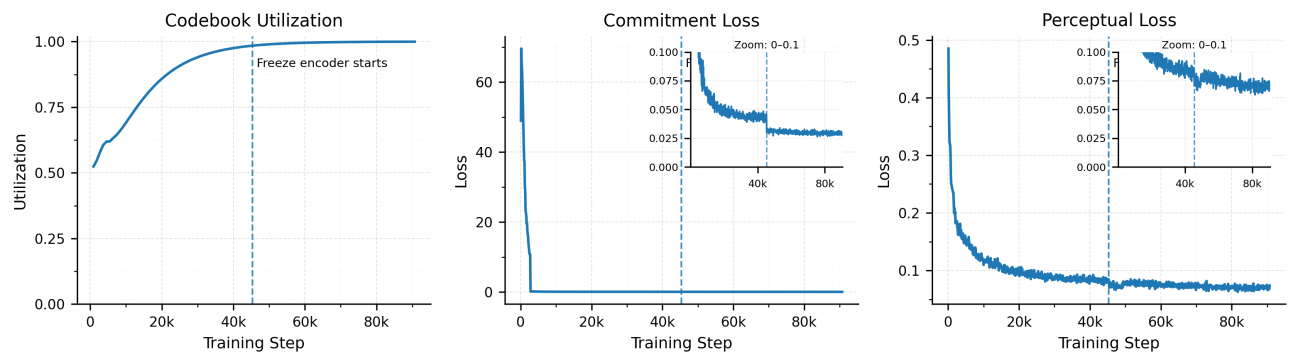}
\caption{Detailed training dynamics on CelebA \(128\times128\), including codebook utilization,
commitment loss, and perceptual loss. Encoder freezing begins at step~45{,}350 in this run, marking
the transition from Stage~1 to the frozen-encoder warm-up stage.}
\label{fig:training-dynamics-detailed}
\end{figure}

Figure~\ref{fig:training-dynamics-detailed} provides detailed CelebA \(128\times128\) training curves
for codebook utilization, commitment loss, and perceptual loss. During Stage~1, codebook
utilization rapidly increases toward nearly full usage, while commitment loss remains relatively high
because the quantizer must adapt to a moving encoder distribution. After the encoder is frozen at
step~45{,}350 in this CelebA run, training enters the warm-up stage and commitment loss decreases
sharply, consistent with the reduced non-stationarity shown in the main text. During warm-up, adversarial training is disabled, which allows the quantizer and decoder to adapt to the fixed encoder manifold using reconstruction-oriented losses. In Stage~2, adversarial training is re-enabled for perceptual refinement.

\begin{figure}[h]
\centering
\includegraphics[width=0.72\linewidth]{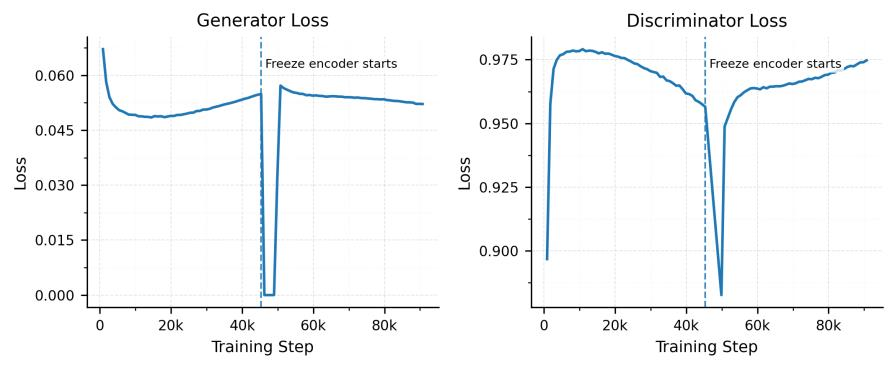}
\caption{Training dynamics of generator and discriminator losses. The adversarial objective is disabled during the frozen-encoder warm-up stage and re-enabled in Stage~2.
}
\label{fig:encoder-drift-detailed}
\end{figure}

\begin{figure}[h]
\centering
\includegraphics[width=0.72\linewidth]{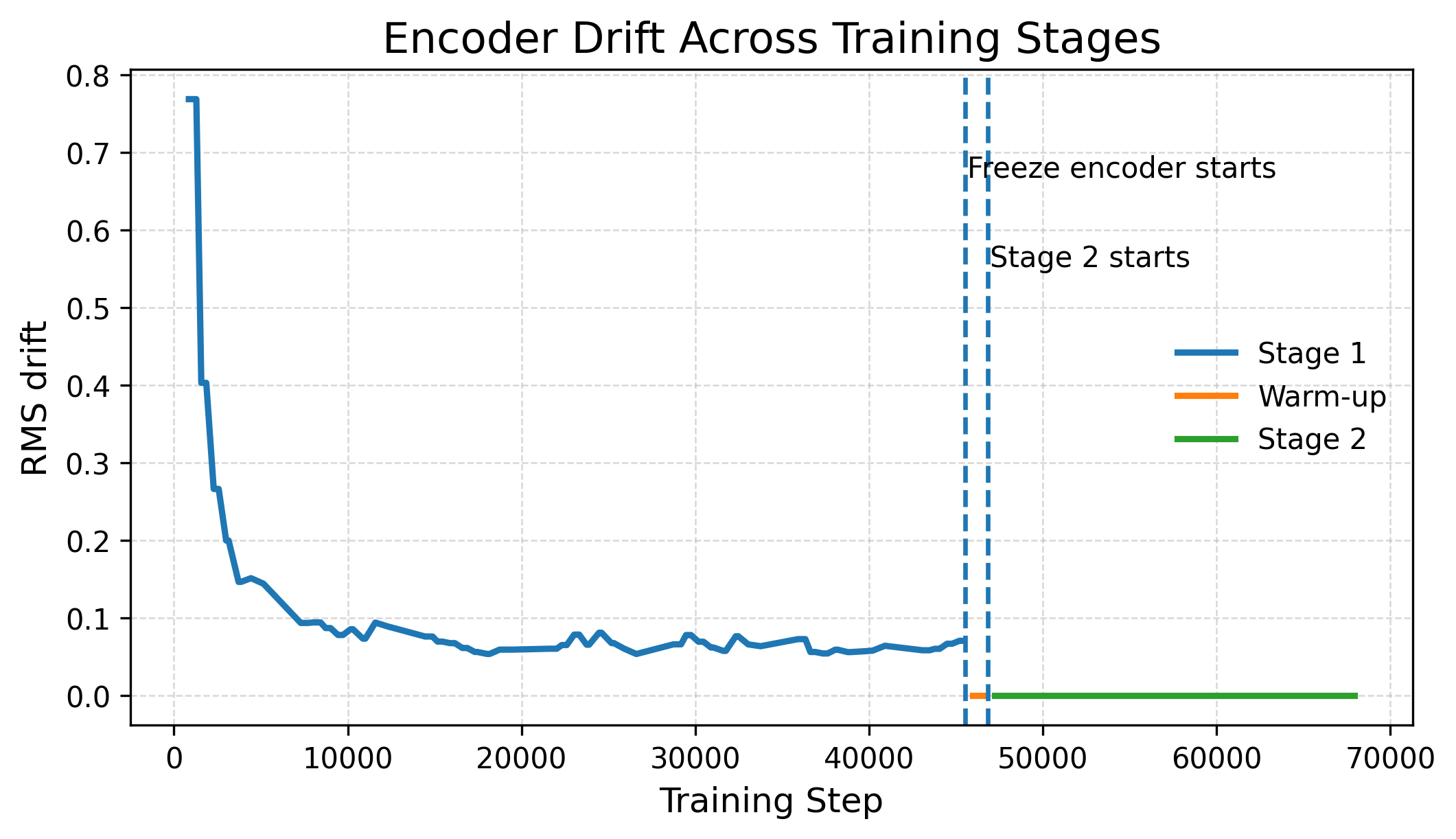}
\caption{Detailed encoder drift across the proposed training phases. Drift is high during Stage~1, drops to approximately zero after the transition to the frozen-encoder warm-up stage, and remains near zero during Stage~2.}
\label{fig:gan-dynamics}
\end{figure}

Figure~\ref{fig:encoder-drift-detailed} shows the adversarial training dynamics. During the warm-up stage, GAN training is disabled to avoid destabilizing the quantizer and decoder immediately after encoder freezing. In Stage 2, adversarial training is reintroduced, and the generator and discriminator losses return to an adversarial optimization process.  Figure~\ref{fig:gan-dynamics} shows the measured encoder drift across the proposed training phases. Drift is large at the beginning of Stage 1, decreases as the encoder approaches convergence, and becomes approximately zero once the encoder is frozen.

\subsection{Computational Cost}
\label{app:computational-cost}

We compare the computational cost of Vanilla VQ, SimVQ, and NSVQ under the same implementation and training environment. As shown in Table~\ref{tab:cost}, NSVQ increases average epoch time from 1055 s to 1257 s relative to SimVQ, corresponding to a $1.20\times$ time overhead. Peak allocated GPU memory increases from 34.47 GB to 38.49 GB, corresponding to a $1.12\times$ memory overhead. Peak reserved memory increases only slightly. These results indicate that the proposed non-stationary dense-update mechanism introduces moderate computational cost.

\begin{table}[h]
\centering
\caption{Computational cost comparison on ImageNet-1k at 128×128 using 4 GPUs and a global batch size of 128. Average epoch time is computed over 3 logged epochs. Peak GPU memory is the maximum observed value. Relative overhead is normalized by SimVQ.}
\label{tab:cost}
\begin{adjustbox}{width=\textwidth}
\begin{tabular}{lccccc}
\toprule
Method & Epoch time & Peak allocated memory & Peak reserved memory & Relative time & Relative allocated memory \\
\midrule
Vanilla VQ & 1048 s & 34.49 GB & 40.75 GB & $1.00\times$ & $1.00\times$ \\
SimVQ & 1055 s & 34.47 GB & 40.77 GB & $1.00\times$ & $1.00\times$ \\
NSVQ (Ours) & 1257 s & 38.49 GB & 41.67 GB & $1.20\times$ & $1.12\times$ \\
\bottomrule
\end{tabular}
\end{adjustbox}
\end{table}

\section{Why Online VQ Converges to a $k$-Means Fixed Point}
\label{app:kmeans}
Let $x\in\mathbb{R}^{d}$ be sampled i.i.d. from a distribution $P$, and let the codebook be $C=(c_1,\ldots,c_K)\in(\mathbb{R}^{d})^K$. Given $C$, define the Voronoi partition
\begin{equation}
V_k(C)=\{x:\|x-c_k\|\leq \|x-c_j\|,\forall j\}.
\end{equation}
The expected quantization distortion is
\begin{equation}
J(C)=\mathbb{E}_{x\sim P}\left[\min_k \|x-c_k\|^2\right]=\sum_{k=1}^{K}\int_{V_k(C)}\|x-c_k\|^2\,dP(x).
\end{equation}
Although the Voronoi boundaries move when the centroids are perturbed, the boundary
terms cancel in the derivative of \(J(C)\). By Leibniz's integral rule, differentiating the
cell integrals gives interior terms and boundary terms. On a shared boundary between
\(V_k(C)\) and \(V_l(C)\), we have
\begin{equation}
\|x-c_k\|_2^2 = \|x-c_l\|_2^2 ,
\end{equation}
so the boundary contributions from adjacent cells cancel. Assuming tie boundaries have
zero probability mass, the gradient is
\begin{equation}
\frac{\partial J}{\partial c_k}(C)
=
-2
\int_{V_k(C)}
(x-c_k)\,dP(x).
\end{equation}
Setting the gradient to zero gives
\begin{equation}
c_k=\mathbb{E}[x\mid x\in V_k(C)],
\end{equation}
which is exactly the fixed-point condition of $k$-means: each centroid equals the conditional mean of its Voronoi cell.

Now consider the standard winner-take-all online VQ update for a sample $x_t$:
\begin{equation}
k_t(x_t)=\arg\min_j \|x_t-c_j^{(t)}\|,
\end{equation}
\begin{equation}
c_k^{(t+1)}=c_k^{(t)}+\eta_t\mathbf{1}[k=k_t(x_t)]\bigl(x_t-c_k^{(t)}\bigr).
\end{equation}
Taking conditional expectation with respect to the current codebook gives
\begin{equation}
\mathbb{E}\left[c_k^{(t+1)}-c_k^{(t)}\mid C^{(t)}\right]=\eta_t h_k(C^{(t)}),
\end{equation}
\begin{equation}
h_k(C)=\mathbb{E}\left[\mathbf{1}[x\in V_k(C)](x-c_k)\right]=\int_{V_k(C)}(x-c_k)\,dP(x).
\end{equation}
Comparing the two expressions shows that $\frac{\partial J}{\partial c_k}(C)=-2h_k(C)$, so the mean online update direction is proportional to $-\nabla J(C)$. Therefore, online VQ is a stochastic approximation to gradient descent on the $k$-means distortion objective.

Equivalently, the recursion can be written as
\begin{equation}
c_k^{(t+1)}=c_k^{(t)}+\eta_t\bigl(h_k(C^{(t)})+\xi_{t,k}\bigr),
\end{equation}
where $\xi_{t,k}$ is a martingale-difference noise term with zero conditional mean.

Under the standard Robbins--Monro conditions $\sum_t\eta_t=\infty$ and $\sum_t\eta_t^2<\infty$, together with bounded second moments and mild regularity assumptions, standard stochastic approximation theory implies that the iterates track the ODE
\begin{equation}
\dot{C}(t)=h(C(t))=-\frac{1}{2}\nabla J(C(t)).
\end{equation}
Using $J(C)$ as a Lyapunov function,
\begin{equation}
\frac{d}{dt}J(C(t))=\langle \nabla J(C(t)),\dot{C}(t)\rangle=-\frac{1}{2}\|\nabla J(C(t))\|^2\leq 0,
\end{equation}
so trajectories converge to the stationary set $\{\nabla J=0\}$. Hence, online VQ converges almost surely to a $k$-means fixed point, typically a local optimum.

\section{Reconstruction Comparison}
\begin{figure}[t]
\centering
\includegraphics[width=0.85\linewidth]{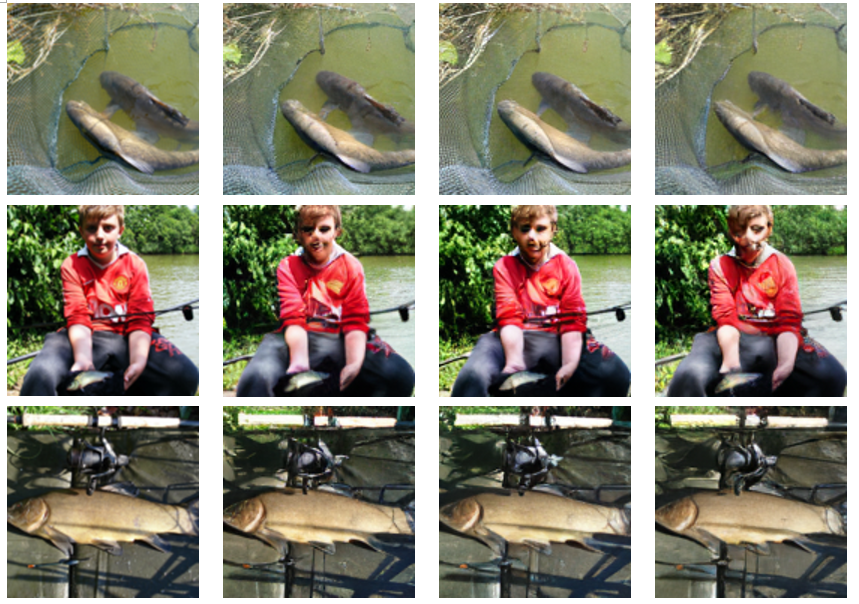}
\caption{Qualitative reconstruction comparison. From left to right: ground truth, NSVQ (our method), SimVQ, and vanilla VQ. Across all three examples, NSVQ produces reconstructions with sharper boundaries, fewer local artifacts, and better structural fidelity than vanilla VQ, while also providing modest visual improvements over SimVQ in fine details and object consistency.}
\label{fig:reconstruction-comparison}
\end{figure}

Figure~\ref{fig:reconstruction-comparison} presents qualitative reconstruction comparisons between the ground truth images, the proposed NSVQ, SimVQ, and vanilla VQ. Consistent with the quantitative results in the main paper, NSVQ produces reconstructions that are generally sharper and structurally more faithful than vanilla VQ, while also showing clearer details than SimVQ in several regions. In the first example, NSVQ better preserves the fish boundaries and the net structure. In the second example, it recovers more coherent facial structure and clothing appearance, whereas vanilla VQ exhibits stronger distortions and SimVQ remains slightly blurrier. In the third example, NSVQ better maintains the elongated body contour of the fish and reduces local artifacts around the supporting structure. Overall, these examples suggest that the proposed training strategy improves not only codebook utilization, but also the perceptual quality and structural consistency of reconstructions.

\section{Implementation Details for Stage Transitions}
To ensure stable and reproducible training, we carefully design the transition between stages to avoid optimization discontinuities or unintended parameter updates. This section summarizes the key implementation details that make the stage-wise training procedure robust in practice.

\paragraph{Optimizer and parameter groups.} Across stages, we maintain consistent optimizer configurations while modifying the set of trainable parameters. When transitioning from Stage~1 to Warm-up stage, the encoder parameters are frozen by setting \texttt{requires\_grad = False}, and the optimizer is updated to exclude these parameters from gradient updates. The decoder and quantizer remain trainable throughout all stages. Importantly, we do not reinitialize the optimizer state when changing stages; instead, we preserve momentum and adaptive statistics, such as Adam moments, for the remaining trainable parameters. This avoids sudden optimization instability caused by restarting the optimizer.

\paragraph{Handling of optimizer state after freezing.} Freezing the encoder effectively removes it from gradient updates but does not invalidate the optimizer state of the remaining modules. Since the encoder no longer contributes gradients, its associated optimizer states become inactive and are ignored. This ensures a smooth transition without requiring explicit state reset or reparameterization.

\paragraph{Gradient control and optimizer toggling.} During adversarial training in Stage~1 and Stage~2, we explicitly control gradient flow using a toggle mechanism for generator and discriminator updates. In our implementation, based on PyTorch Lightning, we ensure that only the generator modules---encoder, decoder, and quantizer---receive gradients during generator updates, and only the discriminator receives gradients during discriminator updates. When entering Warm-up stage, adversarial training is disabled entirely. Both the discriminator updates and the corresponding toggle logic are turned off, so the training reduces to a standard reconstruction objective. This eliminates gradient interference from adversarial objectives and allows the quantizer and decoder to adapt to a fixed latent distribution.

\paragraph{Consistency across stage transitions.} We avoid abrupt changes in optimization dynamics by freezing modules without reinitializing optimizers, gradually reintroducing adversarial training in Stage~2 via a linear ramp, and keeping learning rates and optimizer hyperparameters unchanged across stages. These design choices ensure that stage transitions modify only the intended aspects of training, such as removing encoder drift or reintroducing perceptual refinement, rather than introducing unintended optimization artifacts.

\paragraph{Practical robustness.} In practice, we find that this transition strategy is stable across multiple runs and does not introduce noticeable spikes in loss curves beyond those expected from enabling or disabling adversarial objectives. The observed changes in training dynamics, such as the drop in commitment loss at the beginning of Warm-up stage, are therefore attributable to the intended design of the method, rather than to artifacts of optimizer reinitialization or gradient leakage.

\paragraph{Existing assets and licenses.}
We use ImageNet-1k and CelebA only for academic research evaluation and follow their respective terms of use. ImageNet-1k is used under the ImageNet terms of access, and CelebA is used under its non-commercial research terms. We cite the original dataset papers and websites where applicable. Our implementation builds on standard open-source deep learning libraries such as PyTorch, and evaluation follows commonly used reconstruction metrics including rFID, LPIPS, PSNR, and SSIM. Baseline methods are credited through their original publications, and any released code will include a license file and documentation specifying dependencies and their licenses.